\documentclass[runningheads]{llncs}

\usepackage{eccv}

\usepackage{eccvabbrv}

\usepackage{graphicx}
\usepackage{booktabs}
\usepackage[nolist]{acronym}
\usepackage{multirow}

\usepackage{tikz}
\usetikzlibrary{positioning}
\usetikzlibrary{calc}
\usetikzlibrary{shapes.geometric}
\tikzset{database/.style={cylinder,aspect=0.5,draw,rotate=90,path picture={
\draw (path picture bounding box.160) to[out=180,in=180] (path picture bounding
box.20);
\draw (path picture bounding box.200) to[out=180,in=180] (path picture bounding
box.340);
}}}

\usepackage[dvipsnames]{xcolor}

\newcommand{\imgpathfl}{imgs/small}
\newcommand{\imgpathcatch}{imgs/small}
\newcommand{\imgpathher}{imgs/small}

\usepackage[accsupp]{axessibility}  

\usepackage{hyperref}

\usepackage{orcidlink}

\begin{document}

\title{Style-Extracting Diffusion Models for Semi-Supervised Histopathology Segmentation}

\titlerunning{Style-Extracting Diffusion Models}

\author{Mathias Öttl\inst{1} \and Frauke Wilm\inst{1,4} \and Jana Steenpass\inst{2} \and Jingna Qiu\inst{4} \and Matthias Rübner\inst{3} \and Arndt Hartmann\inst{2} \and Matthias Beckmann\inst{3} \and Peter Fasching\inst{3} \and Andreas Maier\inst{1} \and Ramona Erber\inst{2} \and Bernhard Kainz\inst{4} \and Katharina Breininger\inst{4}}

\authorrunning{M.~Öttl et al.}

\institute{Pattern Recognition Lab, Friedrich-Alexander-Universität Erlangen-Nürnberg (FAU), Germany \and Institute of Pathology, University Hospital Erlangen, FAU, Germany \and Department of Gynecology and Obstetrics, University Hospital Erlangen, FAU, Germany \and Department Artificial Intelligence in Biomedical Engineering, FAU, Germany}

\maketitle

\begin{abstract}

Deep learning-based image generation has seen significant advancements with diffusion models, notably improving the quality of generated images. Despite these developments, generating images with unseen characteristics beneficial for downstream tasks has received limited attention. To bridge this gap, we propose Style-Extracting Diffusion Models, featuring two conditioning mechanisms. 
Specifically, we utilize 1) a style conditioning mechanism which allows to inject style information of previously unseen images during image generation and 2) a content conditioning which can be targeted to a downstream task, e.g., layout for segmentation.
We introduce a trainable style encoder to extract style information from images, and an aggregation block that merges style information from multiple style inputs. This architecture enables the generation of images with unseen styles in a zero-shot manner, by leveraging styles from unseen images, resulting in more diverse generations. In this work, we use the image layout as target condition and first show the capability of our method on a natural image dataset as a proof-of-concept. We further demonstrate its versatility in histopathology, where we combine prior knowledge about tissue composition and unannotated data to create diverse synthetic images with known layouts. This allows us to generate additional synthetic data to train a segmentation network in a semi-supervised fashion. We verify the added value of the generated images by showing improved segmentation results and lower performance variability between patients when synthetic images are included during segmentation training. 
Our code will be made publicly available at [LINK].

\end{abstract}
\begin{figure}[ht]
    \centering
    \resizebox{0.95\textwidth}{!}{%
    \rotatebox[origin=c]{-90}{%
    \begin{tikzpicture}[ image/.style = {inner sep=0pt, outer sep=0pt}, node distance = 1mm and 1mm]
    \def\imageWidth{3cm}
    \def\maskWidth{2.8cm}

    \node[image] (frame1) {\includegraphics[width=\imageWidth]{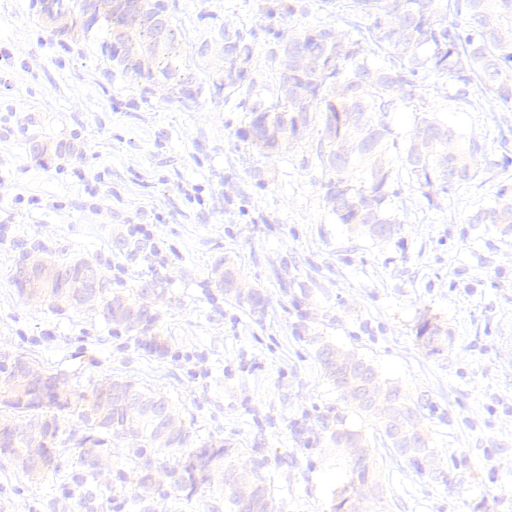}};
    \node[image,below=of frame1] (frame2) {\includegraphics[width=\imageWidth]{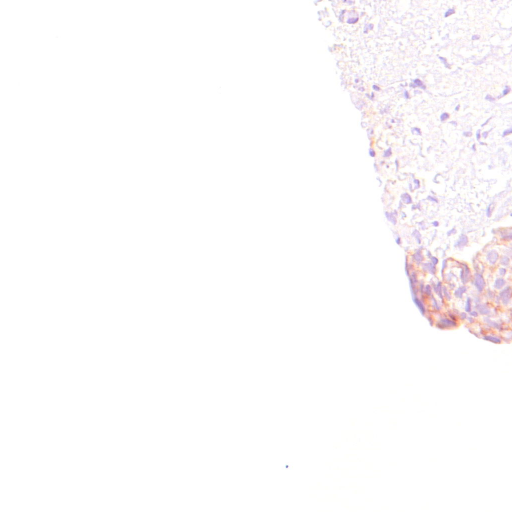}};
    \node[image,below=of frame2] (frame3) {\includegraphics[width=\imageWidth]{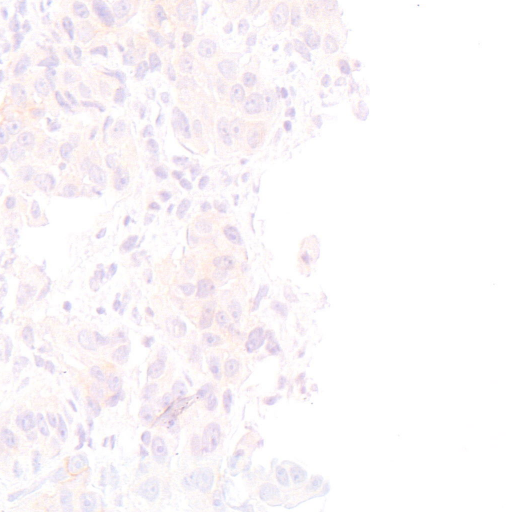}};
    \node[image,below=of frame3] (frame4) {\includegraphics[width=\imageWidth]{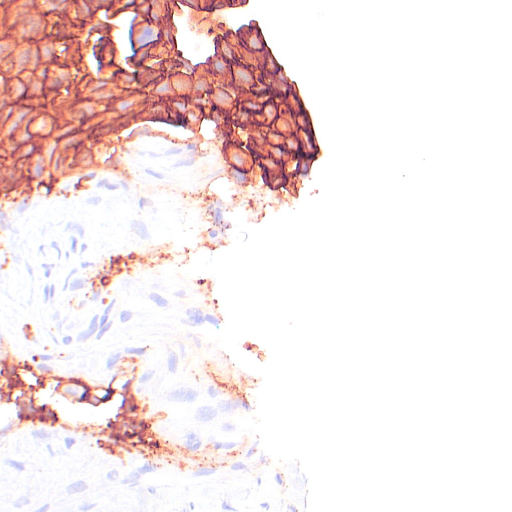}};

    \node [image,right=0.25cm of frame1] (frame5) {\includegraphics[width=\imageWidth]{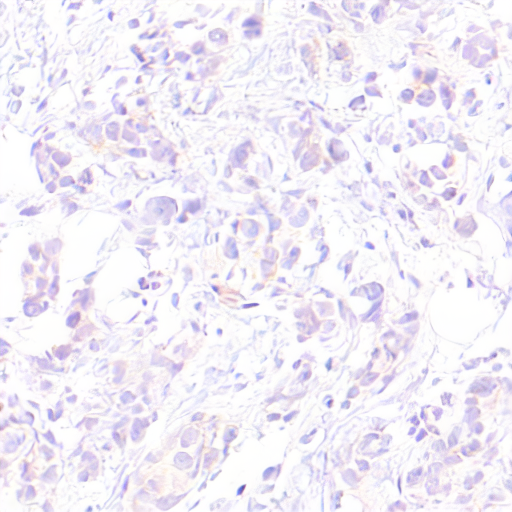}};
    \node [image,below=of frame5] (frame6) {\includegraphics[width=\imageWidth]{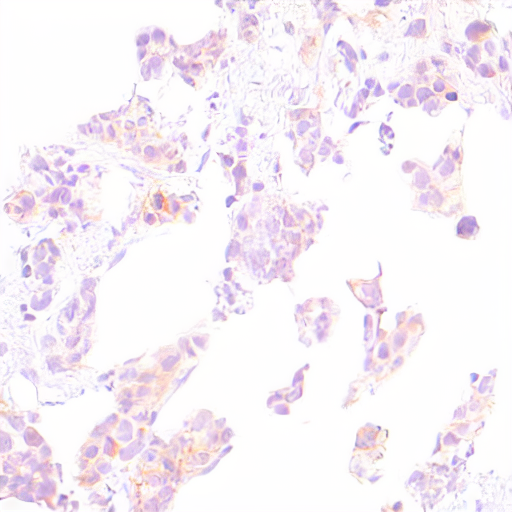}};
    \node [image,below=of frame6] (frame7) {\includegraphics[width=\imageWidth]{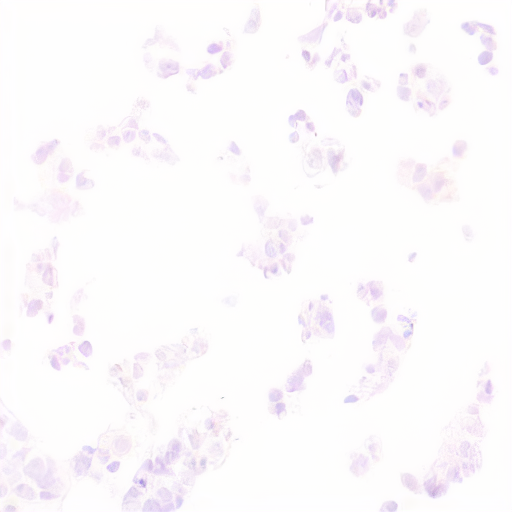}};
    \node [image,below=of frame7] (frame8) {\includegraphics[width=\imageWidth]{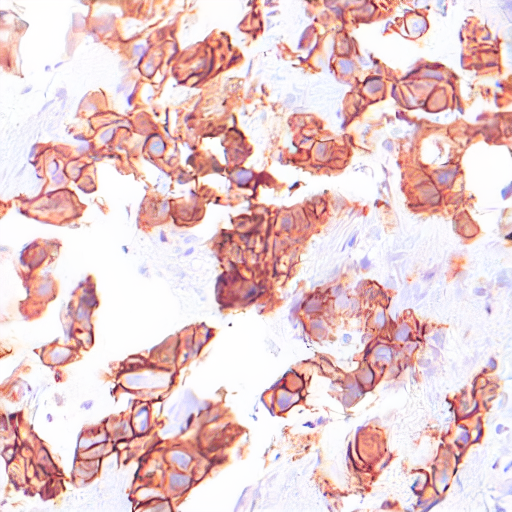}};

    \node[image,above=0.5cm of frame5,draw=black, line width=0.1cm] (maskquery) {\includegraphics[width=\maskWidth]{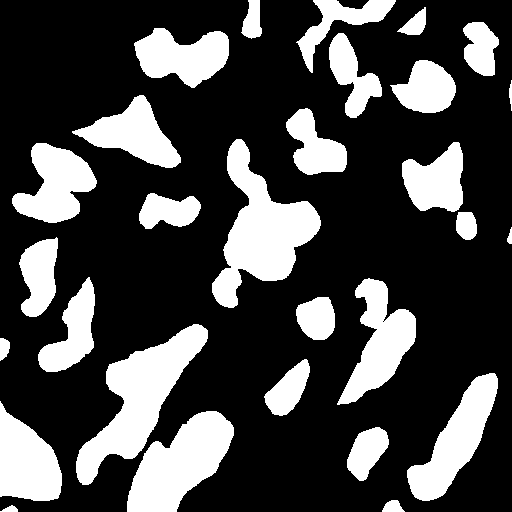}};

    \node[image,right=1cm of frame5,draw=black, line width=0.1cm] (mask1) {\includegraphics[width=\maskWidth]{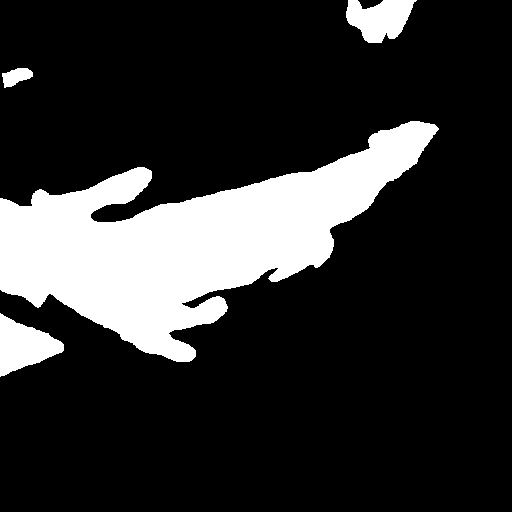}};
    \node[image,right=1cm of frame6,draw=black, line width=0.1cm] (mask2) {\includegraphics[width=\maskWidth]{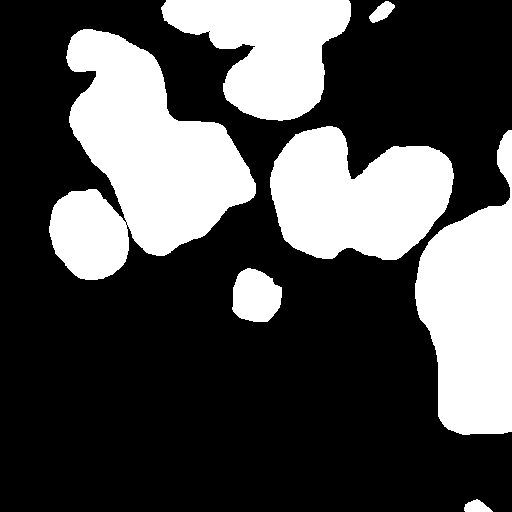}};
    \node[image,right=1cm of frame7,draw=black, line width=0.1cm] (mask3) {\includegraphics[width=\maskWidth]{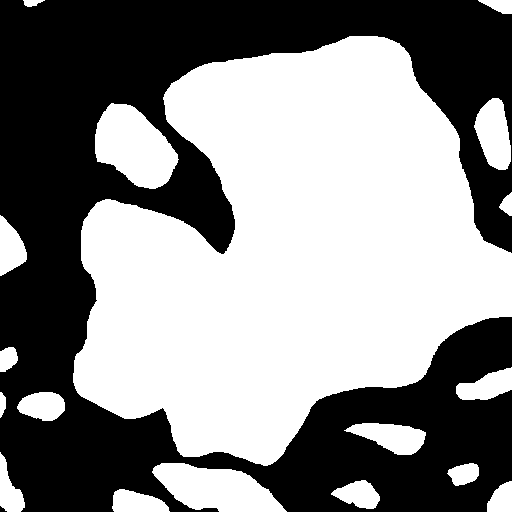}};
    \node[image,right=1cm of frame8,draw=black, line width=0.1cm] (mask4) {\includegraphics[width=\maskWidth]{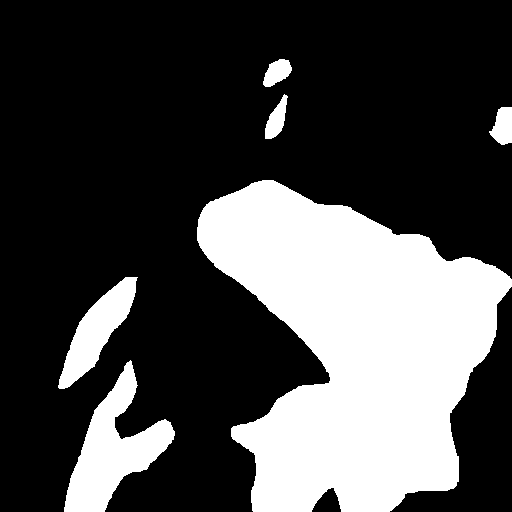}};

    \node[image,right=0.25cm of mask1] (mask5) {\includegraphics[width=\imageWidth]{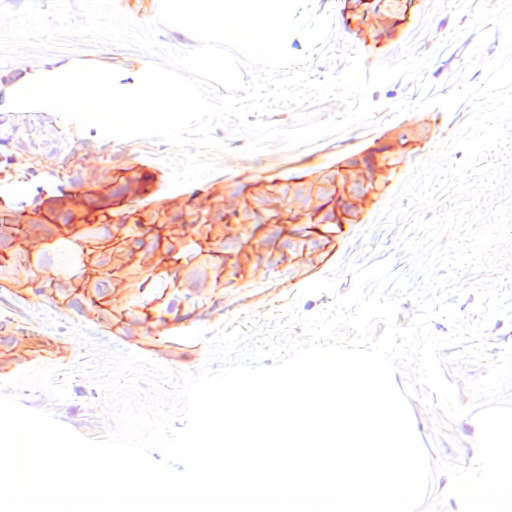}};
    \node[image,below=of mask5] (mask6) {\includegraphics[width=\imageWidth]{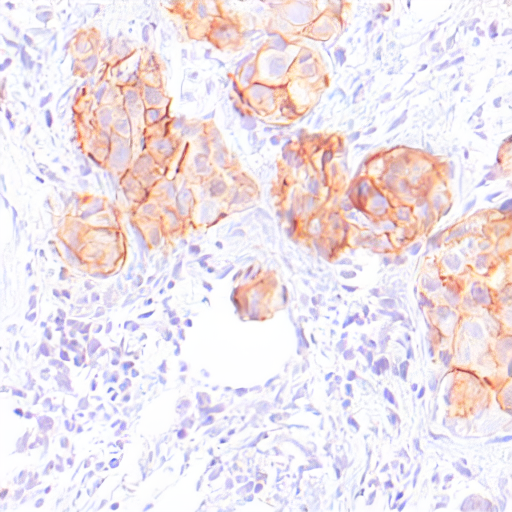}};
    \node[image,below=of mask6] (mask7) {\includegraphics[width=\imageWidth]{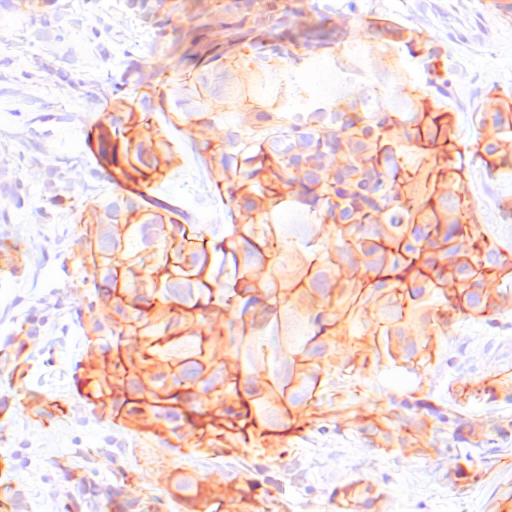}};
    \node[image,below=of mask7] (mask8) {\includegraphics[width=\imageWidth]{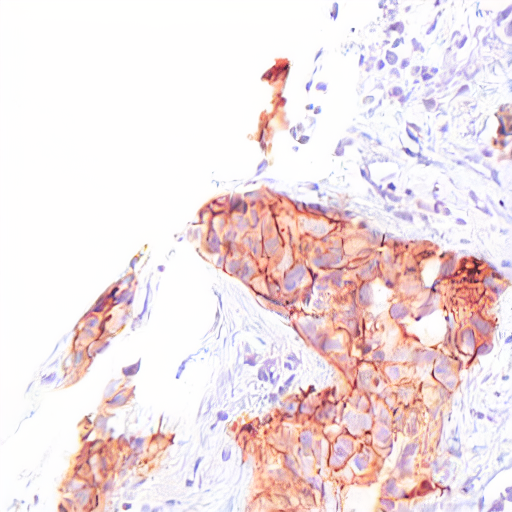}};

    \node[image,above=0.5cm of mask5] (stylequery) {\includegraphics[width=\imageWidth]{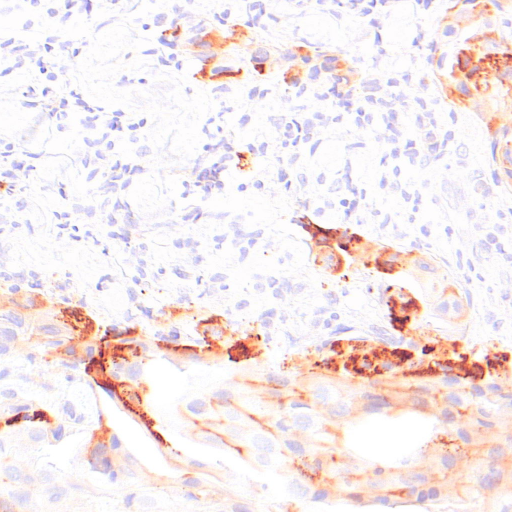}};

    \node[yshift = -2.0cm, inner sep=0] (text1) at ($(frame8)$) {\LARGE{Generated}};
    \node[yshift = -2.0cm, inner sep=0] (text2) at ($(mask8)$) {\LARGE{Generated}};
    \node[yshift = -2.0cm] (text5) at ($(frame4)$) {\LARGE{Styles}};
    \node[yshift = -2.0cm] (text6) at ($(mask4)$) {\LARGE{Layouts}};
    \end{tikzpicture}}
    }
    \caption{Synthetic images with defined layouts and styles generated by our proposed Style-Extracting Diffusion Model (STEDM).}
   \label{fig:teaser}
\end{figure}
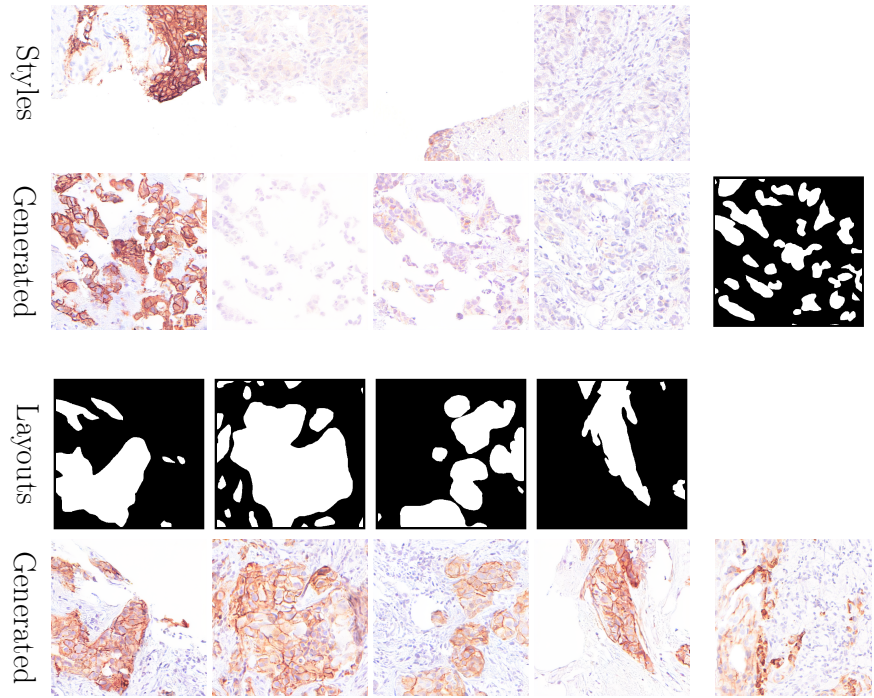

\section{Introduction}
\label{sec:intro}

For many applications, researchers are faced with a small labeled dataset, which can be used for training a supervised task, and a larger unlabelled dataset, which can feature previously unseen objects, concepts, or styles. Integrating these unseen features in the training of the supervised task, e.g., via zero-shot image generation, may improve generalization of the supervised task.
This concept holds particular promise when combined with diffusion models, which have proven to be highly effective in image generation \cite{diff_beat_gan}. Text-to-image diffusion models have shown great success in generating high-quality and diverse images \cite{latent_diff,diff_comp_power}; however, image characteristics are typically learned and controlled by text descriptions, which are limited to concepts that can be expressed verbally and were seen during training. An alternative approach~\cite{few_shot_diff} was proposed, where an encoder extracts consensus information from a set of images to condition a diffusion process. This technique allows the generation of concepts not seen during training. However, without additional constraints, the generated images can suffer from low diversity and offer no value for follow-up tasks.

We introduce \ac{stedm}, featuring simultaneous conditioning on a content conditioning and style information, which is derived from a set of images that specify the desired output style. 
In contrast to prior work, we permit the style information to be a non-linear combination of the style characteristics extracted from the set of images, rather than seeking consensus, thereby facilitating more diverse image generation. 
Our architecture offers the advantage of generating images with a specified content while adopting the style of unseen and potentially unannotated images. Consequently, this approach can harness unannotated data to enhance the diversity of generated outputs.

Due to harnessing the style information of unannotated images, our method holds substantial value for applications where annotations require a high level of expertise and labeling time, e.g., histopathology. Given the large scale of histopathological images, the complexity of labeling, and substantial domain shifts between scanners and patients, our approach is particularly well-suited for this field. 

In this work, we show the potential of our proposed method with semantic layouts as target variable.
We capitalize on prior knowledge of histopathological samples, enabling the training of \ac{stedm} without explicit knowledge about style characteristics. By leveraging unannotated patient data as a source of style information, we can generate images with a predefined semantic layout and unseen styles in a zero-shot manner. The integration of these generated samples into a segmentation training process yields an enhanced and more robust segmentation model, affirming the efficacy of our proposed method.
We summarize the contributions of our work as:
\begin{itemize}
  \item We introduce \acf{stedm}, a novel architectural framework enabling the generation of images with a known content and styles extracted from unseen images.
  \item We demonstrate the applicability of our architecture with the semantic layout as content conditioning on natural images and on histopathology samples, leveraging unique characteristics of histopathological images to train our method and generate images of unseen styles.
  \item We showcase the usability of our approach by integrating the generated images in a semi-supervised fashion to improve semantic segmentation in histopathology as a downstream task, while additionally validating the style diversity of the generated images.
\end{itemize}
\section{Related Work}
\label{sec:related}

\noindent\textbf{Diffusion Models }
have markedly influenced the field of image generation due to their capacity to produce high-quality images. \Acp{ddpm}~\cite{diffusion_ddpm} established the foundation for significant advancements in this domain, leading to models that often surpass the capabilities of \acp{gan}~\cite{diff_beat_gan}.

\Acp{ldm}~\cite{latent_diff} addressed computational complexity issues of diffusion. By reducing the dimensionality of the diffusion process and introducing versatile conditioning mechanisms, \acp{ldm} facilitated more complex and diverse outputs.

Further enhancing the versatility of diffusion models, the few-shot diffusion approach proposed in~\cite{few_shot_diff} utilizes a conditioning vector derived from a set of images using a Vision Transformer (ViT)~\cite{vit}. This technique enables the model to extract a consensus from a set of images, thereby conditioning the diffusion process based on this consensus. This strategy expands the conditioning mechanism from defined labels to information extracted from images.

In the specific context of histopathology~\cite{diff_histo}, diffusion models are starting to show their potential as well. Recent efforts in using diffusion-based image synthesis are noteworthy \cite{her2_diff,cechnicka2023realistic}. However, the application of few-shot or zero-shot diffusion models in histopathology remains largely unexplored.

\noindent\textbf{Image Style Transfer }
involves the separation and recombination of content and style from different images. A first notable approach was proposed in~\cite{baseline_style}, where aesthetic elements of one image are combined with the content of another.

Building on this, \cite{deep_photo} enabled the application of artistic styles to photorealistic images while minimizing distortions common in earlier methods. Similarly, the Swapping Autoencoder for Deep Image Manipulation \cite{style_swapping}, although based on GANs, offered new insights into handling image structure and texture.

In histopathology, style transfer techniques are being explored to address challenges like stain variation in tissue samples. Notable innovations include \cite{stain_style_2,stain_style}. These methods demonstrate the feasibility and effectiveness of applying style transfer in histopathology, predominantly employing GAN-based architectures.

However, a significant limitation in this area is the primary focus on global variations, such as stain or scanner differences, often overlooking patient-specific or tissue variations crucial in histopathological analysis. More adaptable style transfer methodologies are needed to tackle the nuanced requirements in histo\-path\-o\-logy to bridge this gap.
\section{Method}
\label{sec:method}

\begin{figure}[ht]
    \centering
    \resizebox{1.0\textwidth}{!}{%
    \begin{tikzpicture}
        \node[draw=black] (wsi) at (0,0) {\includegraphics[width=3cm]{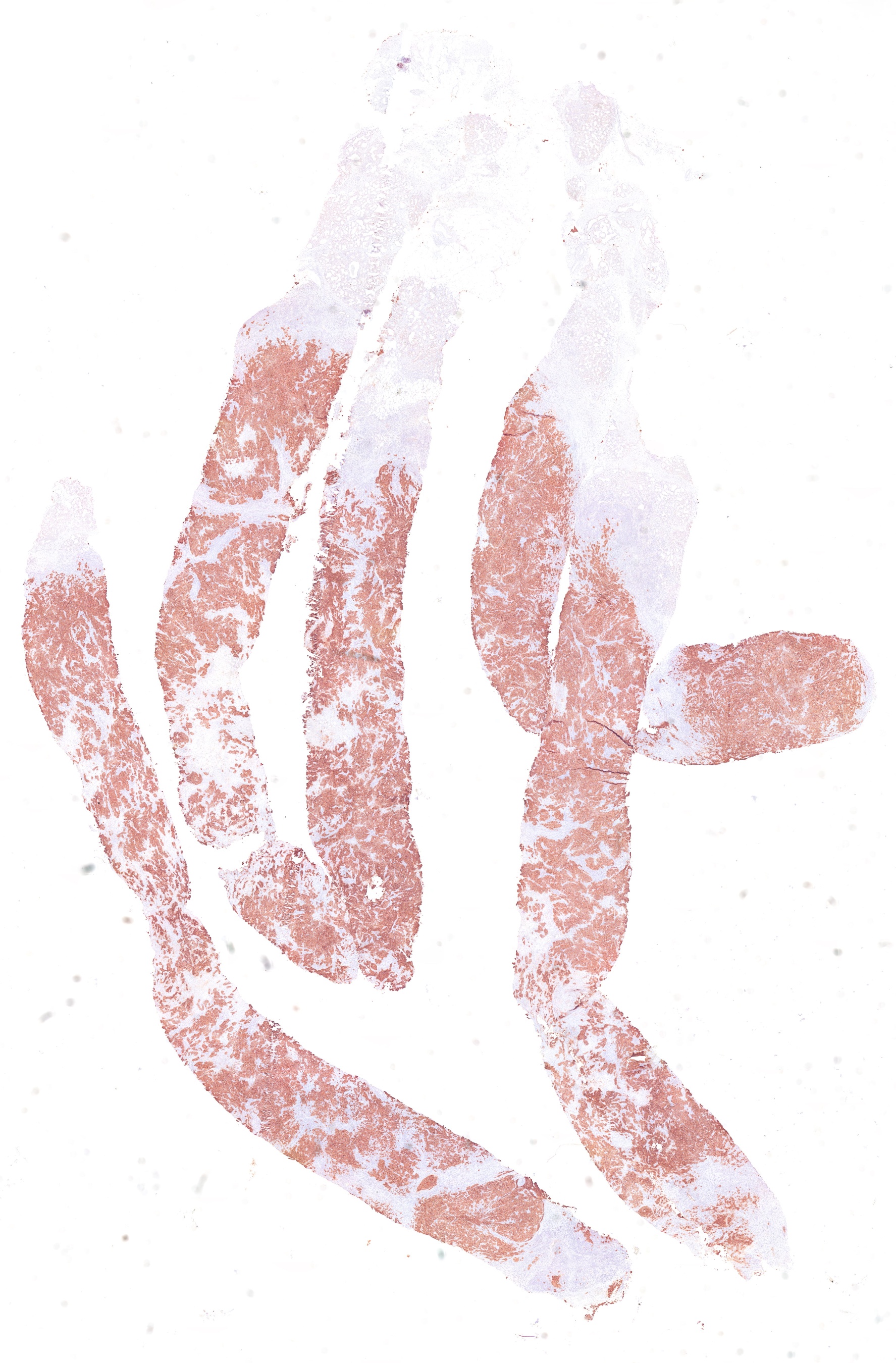}};
        \node (text1) at ($(wsi.south) + (0,-0.5)$) {\Large{WSI}};
        
        \node (image) at ($(wsi.east)+(3,1.25)$) {\includegraphics[width=1.75cm]{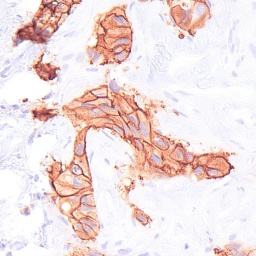}};
        \node (mask) at ($(wsi.east)+(3,-1.25)$) {\includegraphics[width=1.75cm]{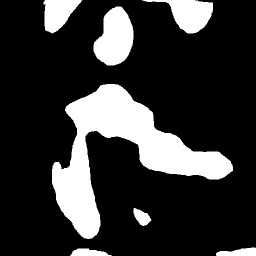}};
        \node[anchor=south] (text-above-img) at (image.north) {\Large{Image $x$}};
        \node[anchor=north] (text-below-mask) at (mask.south) {\Large{Layout $l$}};
        \node[minimum width=2cm] (dummy1) at ($(wsi.east)+(3,0)$) {};

        \node (background) at ($(wsi.east)+(9,0)$) [rounded corners, draw=black, fill=orange!30,minimum width=6.5cm,minimum height=7.5cm] {};
        \node[align=center] at ($(background.north)+(0,0.75)$) {\Large{Style-Extracting} \\ \Large{Diffusion Model (STEDM)}};

        \node (ldm-background) at ($(background.north)+(0,-0.35)$) [rounded corners, draw=black, fill=orange!50, minimum width=5cm, minimum height=3.0cm, anchor=north] {};
        \node[below=0.1cm of ldm-background.north, anchor=north] {\Large{LDM}};

        \draw[shorten >=0.5cm,shorten <=0.25cm,->](wsi.east) -- (dummy1.west) {};
        \draw[shorten >=0.5cm,shorten <=0.25cm,->](dummy1.east) -- (background.west) {};

        \node (rect-l1) at ($(ldm-background)+(-2,0.3)$) [draw,minimum width=0.1cm,minimum height=1.0cm, anchor=south] {};
        \node (rect-r1) at ($(ldm-background)+(2,0.3)$) [draw,minimum width=0.1cm,minimum height=1.0cm, anchor=south] {};
        \node (rect-l2) at ($(ldm-background)+(-1.5,-0.2)$) [draw,minimum width=0.2cm,minimum height=0.6cm, anchor=south] {};
        \node (rect-r2) at ($(ldm-background)+(+1.5,-0.2)$) [draw,minimum width=0.3cm,minimum height=0.6cm, anchor=south] {};
        \node (rect-l3) at ($(ldm-background)+(-1,-0.8)$) [draw,minimum width=0.3cm,minimum height=0.4cm, anchor=south] {};
        \node (rect-r3) at ($(ldm-background)+(+1,-0.8)$) [draw,minimum width=0.3cm,minimum height=0.4cm, anchor=south] {};
        \node (rect-m) at ($(ldm-background)+(0,-1.2)$) [draw,minimum width=1cm,minimum height=0.25cm, anchor=south] {};

        \draw[shorten >=0.25cm,shorten <=0.25cm,->](rect-l1.east) -- (rect-r1.west) {};
        \draw[shorten >=0.25cm,shorten <=0.25cm,->](rect-l2.east) -- (rect-r2.west) {};
        \draw[shorten >=0.25cm,shorten <=0.25cm,->](rect-l3.east) -- (rect-r3.west) {};

        \node [trapezium, trapezium angle=65, minimum width=1.0cm, minimum height=0.9cm, draw, thick] (encoder) at ($(background.south)+(0.0,+0.7)$) {$\mathcal{E}_{s}$};

        \node[inner sep=0, outer sep=0, minimum width=0cm, minimum height=0cm] (dummy2) at ($(wsi.south)+(0,-3.5)$) {};
        \node[draw, rounded corners, align=center, minimum height = 2cm, minimum width = 2.5cm] (text2) at ($(dummy2.east)+(2,0)$) {\Large{Style}\\\Large{Sampling}};
        \draw[shorten >=0.25cm,->]($(text1.south) + (0,-0.25)$) -- (dummy2) -- (text2.west) {};

        \node[anchor=north, draw, rounded corners, minimum height = 1cm] (nearby-block) at ($(text2.south west)+(-0.5,-0.75)$) {\Large{Nearby}};
        \draw[] (text2) -- (nearby-block);
        
        \node[anchor=north, draw, rounded corners, minimum height = 1cm] (multi-patch-block) at ($(text2.south east)+(0.5,-0.75)$) {\Large{Multi-Patch}};
        \draw[] (text2) -- (multi-patch-block);

        \node[xshift=-2cm] (style-image1) at (encoder.south |- text2) {\includegraphics[width=1.75cm]{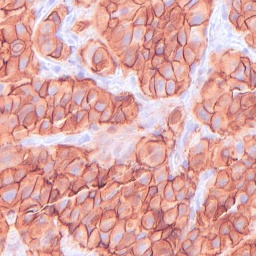}};
        \node[xshift=2cm] (style-image2) at (encoder.south |- text2) {\includegraphics[width=1.75cm]{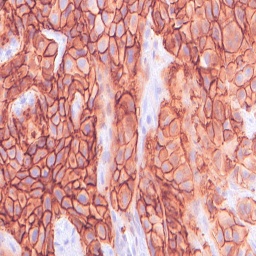}};
        \draw[shorten >=0.25cm,shorten <=0.25cm,->](text2.east) -- (style-image1.west) {};

        \node[yshift=-0.3cm] (label-s1) at (style-image1.south) {\Large{$s_{q,1}$}};
        \node[yshift=-0.3cm] (label-sn) at (style-image2.south) {\Large{$s_{q,n}$}};

        \coordinate (midpoint_style) at ($(style-image1.east)!0.5!(style-image2.west)$);
        \filldraw ($(midpoint_style) + (0.4,0)$) circle (3pt);
        \filldraw ($(midpoint_style) + (0.0,0)$) circle (3pt);
        \filldraw ($(midpoint_style) - (0.4,0)$) circle (3pt);

        \filldraw ($(encoder.north) + (+0.4,0.6)$) circle (3pt);
        \filldraw ($(encoder.north) + (+0.0,0.6)$) circle (3pt);
        \filldraw ($(encoder.north) + (-0.4,0.6)$) circle (3pt);

        \coordinate (features_left) at ($(style-image1 |- encoder.north) + (0, 0.45)$);

        \foreach \i in {1,2,3,4,5} {
            \node [draw, minimum width=0.25cm, minimum height=0.25cm, anchor=south, fill=blue] (rect1-\i) at ($(features_left) + ({\i-3)*0.25},0.0)$) {};
        }
        
        \node[yshift=0.3cm] (vec_1) at (rect1-3.north) {\Large{$v_{s,1}$}};

        \coordinate (features_right) at ($(style-image2 |- encoder.north) + (0, 0.45)$);

        \foreach \i in {1,2,3,4,5} {
            \node [draw, minimum width=0.25cm, minimum height=0.25cm, anchor=south, fill=green] (rect2-\i) at ($(features_right) + ({\i-3)*0.25},0.0)$) {};
        }

        \node[yshift=0.3cm] (vec_2) at (rect2-3.north) {\Large{$v_{s,n}$}};

        \node (agg_block) at ($(encoder.north)+(0.0,+1.2)$) [draw,rounded corners,minimum width=2cm,minimum height=0.65cm, anchor=south] {$agg$};

        \coordinate (features_comb) at ($(agg_block.north) + (0, 0.55)$);

        \node [draw, minimum width=0.25cm, minimum height=0.25cm, anchor=south, fill=green] (rect3-1) at ($(features_comb) + ({1-3)*0.25},0.0)$) {};
        \node [draw, minimum width=0.25cm, minimum height=0.25cm, anchor=south, fill=purple] (rect3-2) at ($(features_comb) + ({2-3)*0.25},0.0)$) {};
        \node [draw, minimum width=0.25cm, minimum height=0.25cm, anchor=south, fill=blue] (rect3-3) at ($(features_comb) + ({3-3)*0.25},0.0)$) {};
        \node [draw, minimum width=0.25cm, minimum height=0.25cm, anchor=south, fill=red] (rect3-4) at ($(features_comb) + ({4-3)*0.25},0.0)$) {};
        \node [draw, minimum width=0.25cm, minimum height=0.25cm, anchor=south, fill=green] (rect3-5) at ($(features_comb) + ({5-3)*0.25},0.0)$) {};

        \node[xshift=0.3cm] (vec_comb) at (rect3-5.east) {\Large{$v_{s}$}};

        \draw[shorten >=0.1cm,shorten <=0.1cm,->](rect3-3.north) -- (rect-m.south) {};
        \draw[shorten >=0.1cm,shorten <=0.1cm,->](agg_block.north) -- (rect3-3.south) {};
        \draw[shorten >=0.1cm,shorten <=0.1cm,->](rect1-3.north east) -- (agg_block.south west) {};
        \draw[shorten >=0.1cm,shorten <=0.1cm,->](rect2-3.north west) -- (agg_block.south east) {};
        \draw[dashed, shorten >=0.1cm,shorten <=0.05cm,->](style-image1.north) -- (rect1-3.south) {};
        \draw[dashed, shorten >=0.1cm,shorten <=0.05cm,->](style-image2.north) -- (rect2-3.south) {};

        \node[anchor= west] (image-out) at ($(wsi.east)+(13.5,0)$) {\includegraphics[width=2cm]{imgs/overview/jpg/train_img}};
        \draw[shorten >=0.25cm,shorten <=0.25cm,->](background.east) -- (image-out.west) {};
        \node[inner sep=0, outer sep=0, minimum width=0cm, minimum height=0cm] (dummy3) at ($(image-out.east)+(1,0)$) {};
        \draw (dummy3 |- background.north) -- (dummy3 |- style-image2.south) {};

        \node[anchor= west] (mask-in) at ($(dummy3)+(2,0)$) {\includegraphics[width=2cm]{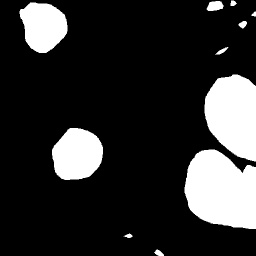}};
        \node[anchor=north] (text-below-mask-pred) at (mask-in.south) {\Large{Known Layout $l$}};
        \node (szsdm) at ($(mask-in.east)+(3,0)$) [align=center, rounded corners, draw=black, fill=orange!30,minimum width=3cm,minimum height=2cm, anchor=west] {\Large{STEDM}};
        \node[anchor= west] (pred-out) at ($(szsdm.east)+(1.5,0)$) {\includegraphics[width=2cm]{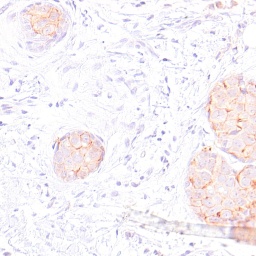}};
        \node[anchor=north] (text-below-img-pred) at (pred-out.south) {\Large{Synthetic Image}};
        \draw[shorten >=0.25cm,shorten <=0.25cm,->](mask-in.east) -- (szsdm.west) {};
        \draw[shorten >=0.25cm,shorten <=0.25cm,->](szsdm.east) -- (pred-out.west) {};

        \node[database, scale=4] (database) at ($(szsdm.north) + (0, 2)$){};
        \node at ($(database) + (0, 1.5)$){\Large{Synthetic Dataset}};
        \node[minimum width=0cm, minimum height=0cm, inner sep=0] (dummy4) at ($(mask-in.north)+(0,2.25)$) {};
        \node[minimum width=0cm, minimum height=0cm, inner sep=0] (dummy5) at ($(pred-out.north)+(0,2.25)$) {};
        \draw[->]($(mask-in.north) + (0,0.25)$) -- (dummy4) -- ($(dummy4)+(4.5,0)$) {};
        \draw[->]($(pred-out.north) + (0,0.25)$) -- (dummy5) -- ($(dummy5)+(-3,0)$) {};

        \node[draw=black, anchor=south] (wsi-pred) at (mask-in |- style-image1.south) {\includegraphics[width=3cm]{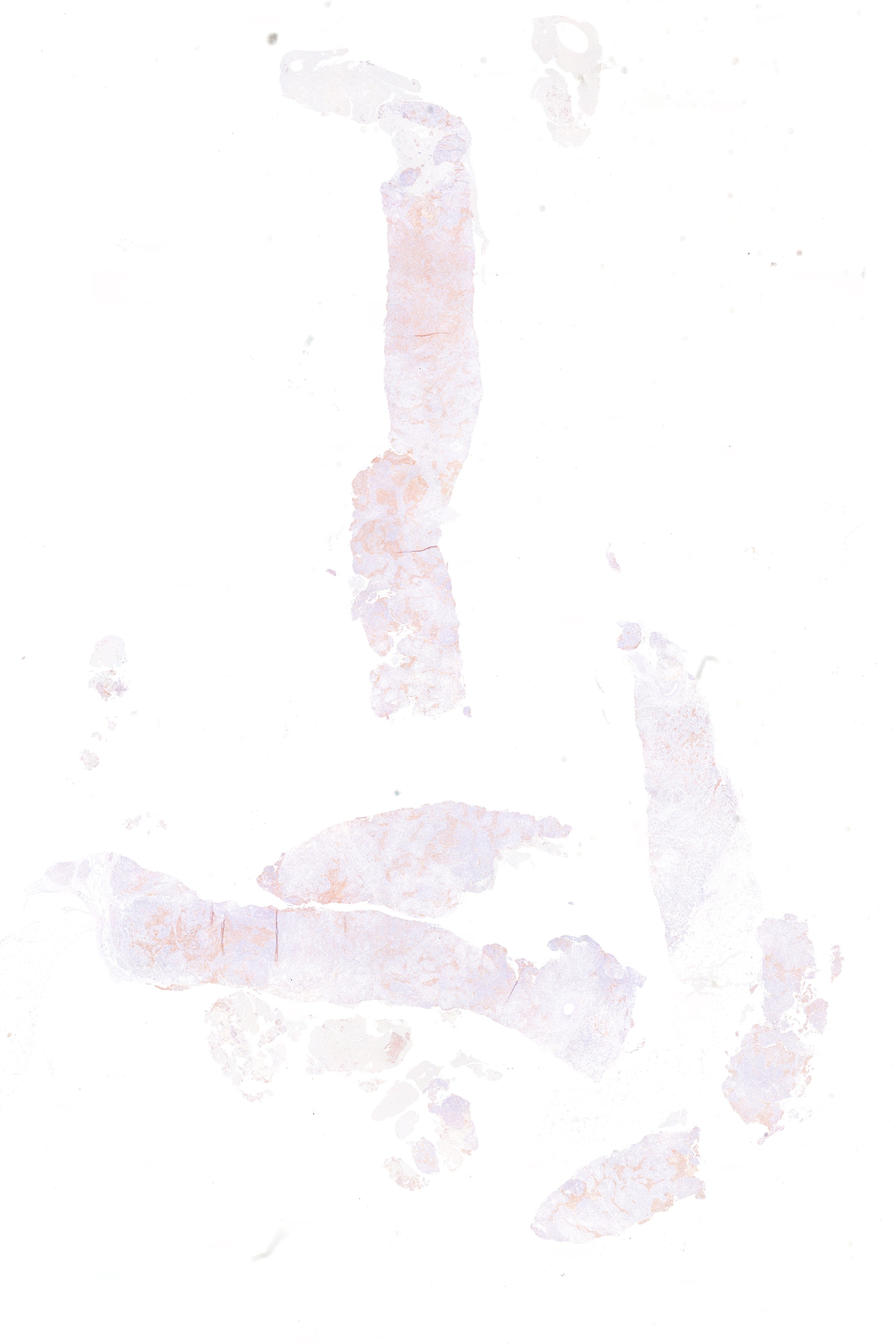}};
        \node (text3) at ($(wsi-pred.south) + (0,-0.5)$) {\Large{Unseen WSI}};
        \node[xshift=-2cm] (style-image3) at (szsdm |- wsi-pred) {\includegraphics[width=2cm]{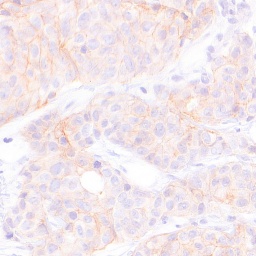}};
        \node[xshift=2cm] (style-image4) at (szsdm |- wsi-pred) {\includegraphics[width=2cm]{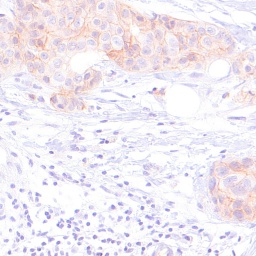}};
        \filldraw ($(style-image3.east) + (0.4,0)$) circle (3pt);
        \filldraw ($(style-image3.east) + (0.8,0)$) circle (3pt);
        \filldraw ($(style-image3.east) + (1.2,0)$) circle (3pt);
        \draw[shorten >=0.25cm,shorten <=0.25cm,->](wsi-pred.east) -- (style-image3.west) {};
        \draw[shorten >=0.25cm,shorten <=0.25cm,->]($(szsdm.south) + (0,-3)$) -- (szsdm.south) {};

        \node[yshift=-0.3cm] (label-s1-p) at (style-image3.south) {\Large{$s_{q,1}$}};
        \node[yshift=-0.3cm] (label-sn-p) at (style-image4.south) {\Large{$s_{q,n}$}};

        \draw[loosely dotted](background.north east) -- (szsdm.north west) {};
        \draw[loosely dotted](background.south east) -- (szsdm.south west) {};

        \node (training) at ($(background.south) +(0,-5)$) {\Huge{Training}};
        \node (inference) at (training -| szsdm) {\Huge{Unseen Style Generation}};

    \end{tikzpicture}}

    \caption{Overview of our proposed \acf{stedm} with the semantic layout as content conditioning for the example of histopathological images during training (left) and inference (right). An image $x$ and corresponding layout query $l$ are sampled, as well as 1 to n style queries $s_{q,n}$. A style encoder $\mathcal{E}_{s}$ extracts a style feature vector $v_{s,n}$ for each style image, which an aggregation block $agg$ combines into a final style feature vector $v_{s}$. An \ac{ldm} is conditioned with the layout query $l$ and the extracted style feature vector $v_{s}$. Synthetic images with unseen styles are generated by taking known layout queries $l$ and combining them with unseen style queries $s_{q,n}$.} 
    \label{fig:overview}
\end{figure}

In this section, we introduce \acf{stedm}, as illustrated in \cref{fig:overview}. At the core of our method is an \ac{ldm} \cite{latent_diff}, which takes the content as first conditioning and style information as second condition. Our approach aims to autonomously extract style information from images, thus eliminating the necessity for predefined style categories and corresponding annotations, and enabling the generation of images with known content and unseen styles. In \cref{sec:sem_dm}, we summarize foundational concepts, including conditional diffusion models, followed by the presentation of \acf{stedm} in \cref{sec:s-zss-dm,sec:zerogen}. Finally, we explore different style sampling strategies for the histopathology use cases in \cref{sec:style_samp}. We adhere to the naming conventions and notation introduced in~\cite{latent_diff}.

\subsection{Conditional Diffusion Model}
\label{sec:sem_dm}

Diffusion models \cite{diff_base,diffusion_ddpm} are latent variable models that learn a given data distribution $p(x)$. The training process is formulated as a Markov Chain with $T$ steps, where a noise estimation model $\epsilon_{\theta}$ is trained to predict the noise $\epsilon$ between a noisy input $x_{t}$ and its one-step denoised version $x_{t-1}$. With uniformly sampled noise steps, the training objective is formulated as
\begin{equation}
    L_{DM}=\mathbb{E}_{x, \epsilon \sim \mathcal{N}(0,1), t}\left[\left\|\epsilon-\epsilon_\theta\left(x_t, t\right)\right\|_2^2\right] \enspace.
\end{equation}

Generating samples with diffusion models requires iteratively applying the noise estimation model $\epsilon_{\theta}$ $T$ times on $x_{t}$, which is computationally expensive, particularly for high-dimensional data like images. Methods to reduce computation load include sampling strategies \cite{diff_ddim,diff_dpmpp} and \acp{ldm}~\cite{latent_diff}, which operate on a latent representation $z$. Here an encoder $\mathcal{E}$ extracts $z=\mathcal{E}(x)$, and a decoder $\mathcal{D}$ restores $x=\mathcal{D}(z)$. The training objective is
\begin{equation}
    L_{LDM}=\mathbb{E}_{\mathcal{E}(x), \epsilon \sim \mathcal{N}(0,1), t}\left[\left\|\epsilon-\epsilon_\theta\left(z_t, t\right)\right\|_2^2\right] \enspace .
\end{equation}

Diffusion models can not only model data distributions $p(x)$ but also conditional data distributions $p(x|y)$ \cite{diff_beat_gan}.
This results in the training objective for conditional latent diffusion models being
\begin{equation}
    L_{CLDM}=\mathbb{E}_{\mathcal{E}(x), \epsilon \sim \mathcal{N}(0,1), t}\left[\left\|\epsilon-\epsilon_\theta\left(z_t, t, y\right)\right\|_2^2\right] \enspace .
\end{equation}

\subsection{Style-Extracting Diffusion Model}
\label{sec:s-zss-dm}

Prior works introduced the idea of images being composed of both style and content \cite{baseline_style,style_swapping}. Building on this concept, we define that each image $x$ contains content information $c$ and style information $s$, represented as $x = \{c,s\}$.

In conditional diffusion models, the content information $c$ is supplied through a content query $c$ to the network, while the model unconditionally infers the style information $s$. Consequently, during sample generation, the content can be queried, while the model selects the style from the training data.

To gain control over the style of generated samples, we model our data distribution to also be conditional on the style information $s$, resulting in $p(x|c,s)$. However, incorporating style information $s$ into the diffusion model for training poses a challenge due to potential variations in style information across different levels of detail or positions within an image.

Motivated by~\cite{few_shot_diff}, we instead condition the diffusion model with a trainable style encoder $\mathcal{E}_{style}$. This encoder takes a style query image $s_{q}$ and extracts a vector $v_{s}$ containing only the style information, disregarding the layout:
\begin{equation}
    \mathcal{E}_{style}(s_{q})=v_{s} \enspace .
    \label{style}
\end{equation}

To address scenarios where a single style query image $s_{q}$ may not contain the entire style information $s$, we propose employing $1$ to $n$ style query images $s_{q,n}$. To consolidate the style vectors $v_{s,n}$ extracted from each style query image $s_{q,n}$ into one combined style vector $v_{s}$, we introduce an aggregation block $agg$, described by
\begin{equation}
    v_{s} = agg(v_{s,1},...,v_{s,n}).
    \label{agg}
\end{equation}

The motivation for the aggregation block $agg$ over a simple averaging of the style vectors $v_{s,n}$ is to enable a non-linear combination, for cases where the style information needs to be assembled non-uniformly from multiple images. We incorporate the style encoder $\mathcal{E}_{style}$ and the aggregation block $agg$ into the training procedure, enabling the model to learn both, style feature extraction and the aggregation over multiple style query images.

This results in the training objective of our proposed method, with $v_{s}$ defined in \cref{agg,style}:
\begin{equation}
    L_{STE}=\mathbb{E}_{\mathcal{E}(x), \epsilon \sim \mathcal{N}(0,1), t}\left[\left\|\epsilon-\epsilon_\theta\left(z_t, t, c, v_{s}\right)\right\|_2^2\right]
\end{equation}

In this work, we will demonstrate the concept and feasibility of our proposed method with semantic layouts as content queries; we note, however, that other types of conditions, e.g., class labels or text, would work similarly.


\subsection{Unseen Style Image Generation}
\label{sec:zerogen}

We introduce style conditioning through style query images, eliminating the need to define and annotate style categories, thereby offering greater freedom and flexibility. During image generation, we can pair arbitrary content queries $c$ with style query images $s_{q,1,...,n}$ from unseen data, potentially containing style variations not present in the training data. 
This enables generation of images with predefined content while incorporating style information from unseen images.
Utilizing classifier-free guidance\cite{diff_cfg} during training and inference allows us to put additional emphasis on styles not seen during training, resulting in a zero-shot style generation.
Since we do not need to know the content of the style images, we can leverage large amounts of unannotated data to create diverse synthetic images. The generated images can contain additional variability, which could have potential utility for subsequent tasks, such as semi-supervised segmentation.

\subsection{Style Sampling}
\label{sec:style_samp}

In our proposed approach, the style information $s$ is derived from the style query images $s_{q,1,...,n}$ using a trainable style encoder $\mathcal{E}_{style}$ and an aggregation block $agg$. Since we do not provide a predefined style label, the nature of the extracted style information depends on the chosen style query images $s_{q,1,...,n}$. During training, the style encoder $\mathcal{E}_{style}$ learns to detect and extract the shared information between the image $x$ and the style query images $s_{q,1,...,n}$. What kind of style information the encoder $\mathcal{E}_{style}$ can infer depends on the data and sampling strategy.

The simplest style sampling strategy involves sampling augmented versions of the image $x$ as style query images $s_{q,1,...,n}$. To ensure that the style encoder $\mathcal{E}_{style}$ learns to capture only the style information $s$, alterations have to be performed to the content information $c_{s}$ of the style query image $s_{q}$, so that it does not match the content information $c_{x}$ of the input image $x$. Otherwise, the style encoder $\mathcal{E}_{style}$ might also capture content information $c$, leading to a bias between content information $c$ and style information $s$. For this style sampling strategy, a single style query image $s_{q}$ suffices, as the complete style information should be present in one augmented copy of the input image $x$.

In this work, we consider histopathology images as one use case, where we recognize the presence of multiple levels of style information, such as scanner characteristics, patient characteristics, and tissue characteristics. Due to large size of histopathology images, patch-based processing is typically used. While scanner characteristics are typically captured in any style patch from the same patient, capturing local characteristics poses a greater challenge. We propose two style sampling strategies for histopathology images.

First, we suggest a nearby style sampling strategy, where the style query images $s_{q,1,...,n}$ are sampled from a spatially close location to the image $x$. This strategy aligns with the biological properties of tissue, where local areas often exhibit homogeneous characteristics. With this approach, we do not need to manually define augmentations to the content information $c$, because the sampled style image should contain 
the same style information $s$ due to the spatial proximity but 
a different content (e.g., a different arrangement of cells). We assume a single style query image to be sufficient for this sampling strategy. During image generation, we have to sample the style query image $s_{q,1}$ from random locations.

Second, we propose a multi-patch-based sampling strategy, where multiple style query images $s_{q,1,...,n}$ are randomly sampled from the tissue of the current patient to capture (all) different tissue characteristics.
The required number $n$ of style query images depends on the diversity of the data, such as the ratio between tumor and non-tumor tissue. For this style sampling strategy, we consider the aggregation block $agg$ crucial, as the style information in the image $x$ may be distributed among the style query images $s_{q,1,...,n}$. For instance, one image might contain relevant information about the background tissue, while another contains the style information about the tumor tissue. For image generation, we can sample the style query images $s_{q,1,...,n}$ with the same strategy as during training.
\section{Experiments and Results}
\label{sec:experiments_results}

\subsection{Datasets}
\label{sub_sec:datasets}

\textbf{Flower Dataset.} For proof of principle, we investigate our proposed approach on the Oxford 102 flowers dataset~\cite{flowers_1,flowers_2}. Comprising 102 flower classes found in the United Kingdom, the dataset contains 40-258 images per category, totaling 8189 images and segmentations. To investigate the method's ability to capture styles unseen or underrepresented during training, we manually excluded flower classes primarily featuring the colors blue, purple, and pink, resulting in the exclusion of 54 classes. The class splits, including examples, are shown in the supplementary material.
To simulate different amounts of annotated data, we select 960, 480 or 144 images as training data, equally distributed among the remaining 48 classes. The images excluded from training serve as style source.

\noindent\textbf{HER2 Dataset.} The first histopathological dataset used in this work consists of tissue sections from breast cancer, immunohistochemically stained for \ac{her2} expression \cite{her2_base}, with 600 \acp{wsi} from as many patients, digitized using a 20x objective on a PANNORAMIC 1000 scanner from 3DHistech. For 40 patients, manual segmentation annotations are available for twelve \ac{roi} each.
A medical student created manual segmentation annotations of tumor tissue for 32 patients for training and validation, while a pathologist annotated the remaining eight for testing. The 560 patients without annotations serve as a style source during image generation. To test the method with varying amounts of data, we use 24, 12, or 4 annotated patients.
The data is split stratified according to known tumor subtype variations (equal distribution of \ac{her2} scores), leaving patients-wise and local tissue style variations as main variations tackled in this work.

\noindent\textbf{CATCH Dataset.} We further use the publicly available \acf{catch} by Wilm \etal~\cite{Wilm2022}. With 350 \acp{wsi} from 282 individual canine patients, the H\&E-stained tissue slides were digitized using Leica ScanScope CS2 linear scanners, at a resolution of $0.25\,\mu m$ per pixel. In this dataset, seven different tumor types are included, with 50 \acp{wsi} each. Annotation of tissue segmentation maps was performed by a pathologist, as well as three medical students under a pathologist supervision. We consider all tumor subtypes as a combined tumor class and the rest of the areas as a background class. Using 42, 21, and 7 \ac{wsi}, equally distributed between known tumor types for training. The remaining \acp{wsi} are considered as style source, with patient-wise style variations being assumed the most relevant variation.

\subsection{Implementation}

\begin{figure*}
    \centering
    \resizebox{1.0\textwidth}{!}{%
    \begin{tikzpicture}[ image/.style = {inner sep=0pt, outer sep=0pt}, node distance = 1mm and 1mm]
    \def\imageWidth{3cm}
    \def\maskWidth{2.8cm}
    
    \node [image] (style1) {\includegraphics[width=\imageWidth]{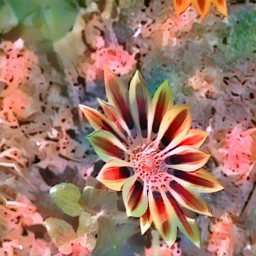}};
    \node [image,right=of style1] (style2) {\includegraphics[width=\imageWidth]{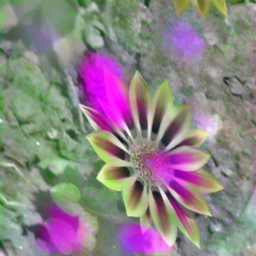}};
    \node [image,right=of style2] (style3) {\includegraphics[width=\imageWidth]{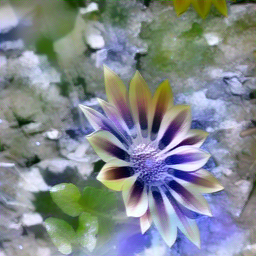}};

    \node[image,below=of style1] (style4) {\includegraphics[width=\imageWidth]{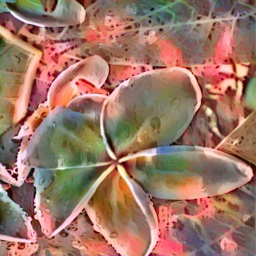}};
    \node[image,right=of style4] (style5) {\includegraphics[width=\imageWidth]{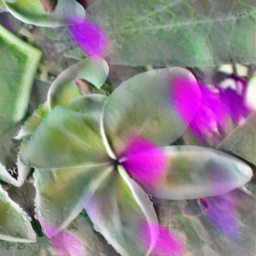}};
    \node[image,right=of style5] (style6) {\includegraphics[width=\imageWidth]{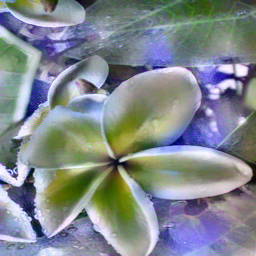}};

    \node[image,below=of style4] (style7) {\includegraphics[width=\imageWidth]{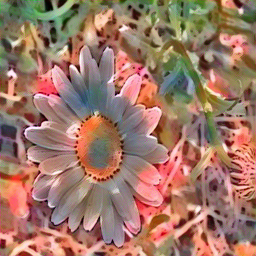}};
    \node[image,right=of style7] (style8) {\includegraphics[width=\imageWidth]{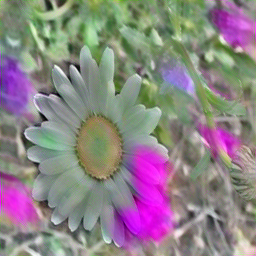}};
    \node[image,right=of style8] (style9) {\includegraphics[width=\imageWidth]{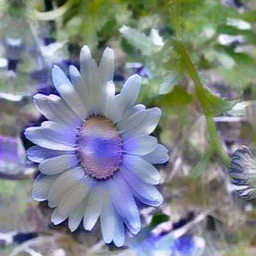}};

    \node [image, right=1cm of style3] (diff1) {\includegraphics[width=\imageWidth]{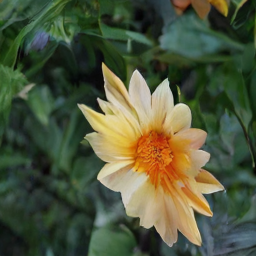}};
    \node [image,right=of diff1] (diff2) {\includegraphics[width=\imageWidth]{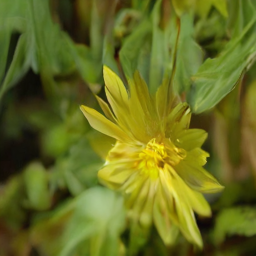}};
    \node [image,right=of diff2] (diff3) {\includegraphics[width=\imageWidth]{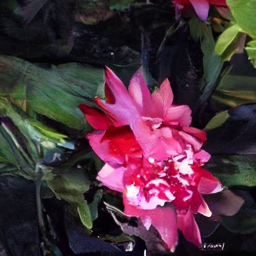}};

    \node[image,below=of diff1] (diff4) {\includegraphics[width=\imageWidth]{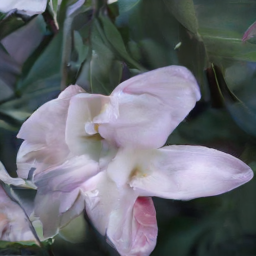}};
    \node[image,right=of diff4] (diff5) {\includegraphics[width=\imageWidth]{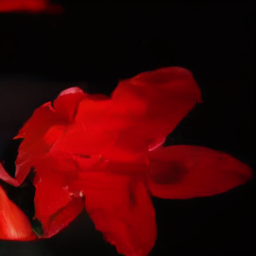}};
    \node[image,right=of diff5] (diff6) {\includegraphics[width=\imageWidth]{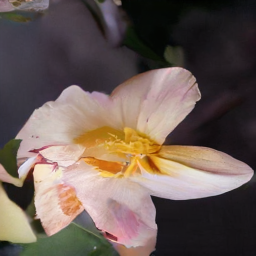}};

    \node[image,below=of diff4] (diff7) {\includegraphics[width=\imageWidth]{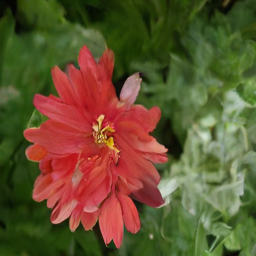}};
    \node[image,right=of diff7] (diff8) {\includegraphics[width=\imageWidth]{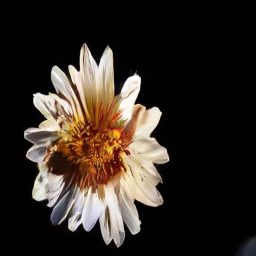}};
    \node[image,right=of diff8] (diff9) {\includegraphics[width=\imageWidth]{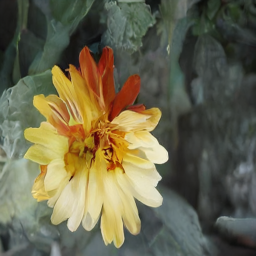}};

    \node [image,right=1cm of diff3] (nearby1) {\includegraphics[width=\imageWidth]{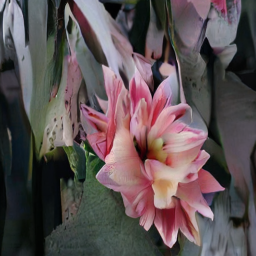}};
    \node [image,right=of nearby1] (nearby2) {\includegraphics[width=\imageWidth]{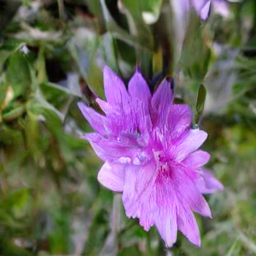}};
    \node [image,right=of nearby2] (nearby3) {\includegraphics[width=\imageWidth]{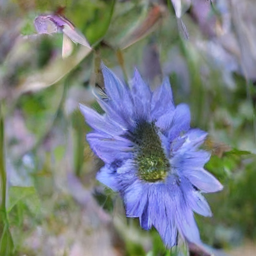}};

    \node[image,below=of nearby1] (nearby4) {\includegraphics[width=\imageWidth]{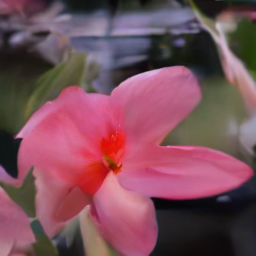}};
    \node[image,right=of nearby4] (nearby5) {\includegraphics[width=\imageWidth]{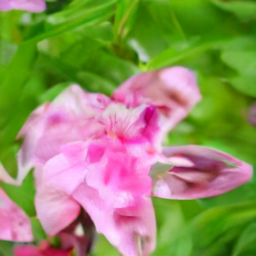}};
    \node[image,right=of nearby5] (nearby6) {\includegraphics[width=\imageWidth]{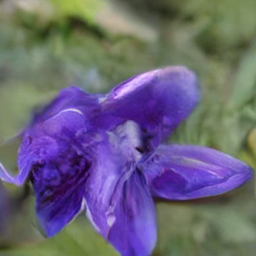}};

    \node[image,below=of nearby4] (nearby7) {\includegraphics[width=\imageWidth]{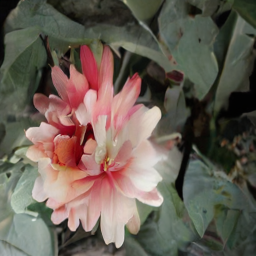}};
    \node[image,right=of nearby7] (nearby8) {\includegraphics[width=\imageWidth]{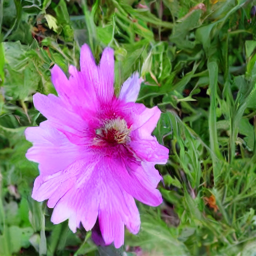}};
    \node[image,right=of nearby8] (nearby9) {\includegraphics[width=\imageWidth]{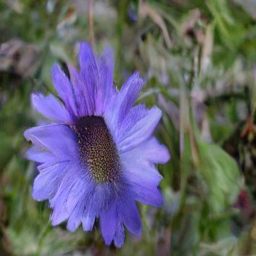}};

    \node[image,left=0.5cm of style1,draw=black, line width=0.1cm] (mask1) {\includegraphics[width=\maskWidth]{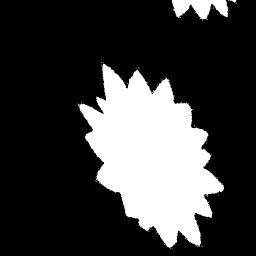}};
    \node[image,left=0.5cm of style4,draw=black, line width=0.1cm] (mask2) {\includegraphics[width=\maskWidth]{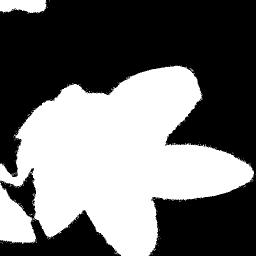}};
    \node[image,left=0.5cm of style7,draw=black, line width=0.1cm] (mask3) {\includegraphics[width=\maskWidth]{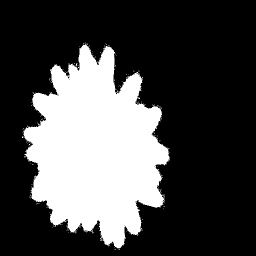}};

    \node[image,above=1cm of style1] (style_q1) {\includegraphics[width=\imageWidth]{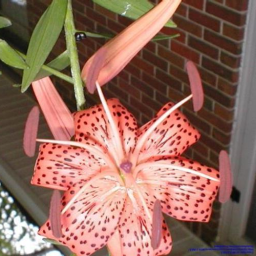}};
    \node[image,above=1cm of style2] (style_q2) {\includegraphics[width=\imageWidth]{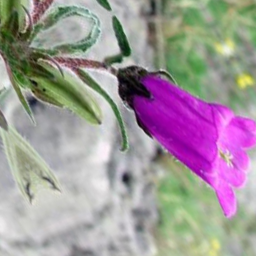}};
    \node[image,above=1cm of style3] (style_q3) {\includegraphics[width=\imageWidth]{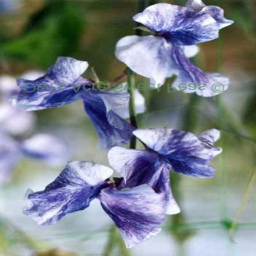}};

    \node[image,above=1cm of nearby1] (style_q4) {\includegraphics[width=\imageWidth]{\imgpathfl/flowers_comb/styl_img_0.png}};
    \node[image,above=1cm of nearby2] (style_q5) {\includegraphics[width=\imageWidth]{\imgpathfl/flowers_comb/styl_img_1.png}};
    \node[image,above=1cm of nearby3] (style_q6) {\includegraphics[width=\imageWidth]{\imgpathfl/flowers_comb/styl_img_3.png}};

    \node[yshift = -2.0cm, inner sep=0] (text1) at ($(style7)!0.5!(style9)$) {\LARGE{Style Transfer~\cite{baseline_style}}};
    \node[yshift = -2.0cm, inner sep=0] (text1) at ($(diff7)!0.5!(diff9)$) {\LARGE{Semantic Diffusion}};
    \node[yshift = -2.0cm, inner sep=0] (text1) at ($(nearby7)!0.5!(nearby9)$) {\LARGE{Ours (Augmented Style Cond.)}};
    
    \node[inner sep=0] (text2) at (mask3|- text1) {\LARGE{Layouts}};
    \node[yshift = 2.0cm] (text3) at ($(style_q4)!0.5!(style_q6)$) {\LARGE{Styles}};
    \node[yshift = 2.0cm] (text3) at ($(style_q1)!0.5!(style_q3)$) {\LARGE{Styles}};
    \end{tikzpicture}}
    \caption{Image generation results with the flower dataset, for the style transfer baseline~\cite{baseline_style} (left), a semantic conditioned diffusion model (center) and our proposed method trained with augmented images as style source (right). Our method is able to generate flowers with colors that were absent or underrepresented in the training data.}
    \label{fig:flowers_gen}
\end{figure*}

We implemented our method for the case of semantic layouts as content conditioning.
Our implementation is based on the publicly available implementation of latent diffusion \cite{latent_diff}, using settings and pre-trained weights for landscape synthesis ($512^2$ finetuned), including the pre-trained VQ-F4 autoencoder. The style encoder $\mathcal{E}_{style}$ is a Swin V2 Tiny Transformer \cite{swin_v2}, and the aggregation block $agg$ is implemented as two linear layers with ReLU activations. The extracted style vector $v_{s}$ is fed into an embedding block at the bottleneck of the \ac{ldm}. The semantic layouts are downsampled to match the output size of the VQ-F4 autoencoder and are concatenated to the input latent representation.

All experiments used $512^2$ resolution images. Augmented style query images were generated by applying affine transformations on the sampled image. For both histopathological datasets, patches of $512^2$ were sampled with the OpenSlide library. For nearby style sampling, a single patch was extracted within 512 pixels of the sampled image. Multi-patch sampling involved sampling ten patches from random locations within the tissue area, which was detected with thresholding. We randomly dropped the style query images $s_{q,1,...,n}$ for 25\% of the elements to enable classifier-free guidance \cite{diff_cfg}. Models were trained for 25 epochs, with 10,000 samples each.

Image generation was performed with 128 steps of DDIM sampling~\cite{diff_ddim} and a 1.5 classifier-free guidance scale, which was chosen based on a visual assessment of generated images. Example images for different classifier-free guidance scales are shown in the supplementary material. Semantic layout queries $l_{q}$ were sampled from the annotated data and augmented with affine transformations. Style queries were sampled from random locations within the tissue area. A total of 20000 synthetic images and corresponding masks were created to be used in the semi-supervised segmentation experiments. 
As first style transfer baseline, we utilized the PyTorch implementation of ``A Neural Algorithm of Artistic Style''~\cite{baseline_style}. The algorithm ran for 300 epochs with a style weight of $10^6$, on content images from the annotated data and unseen style images. As second style transfer baseline we included Swapping Autoencoders\cite{style_swapping}. We trained with the settings provided for the LSUN church dataset. For generation we sampled images with known layout from the validation data and applied the style transfer with images from the unlabeled data.
As plain diffusion model baseline we employ a semantic conditioned diffusion model, where no style conditioning is provided.
To validate the effect of our style encoder $\mathcal{E}_{style}$ and aggregation block $agg$ we perform experiments for the histopathology datasets with the sVit~\cite{few_shot_diff} as style extractor. We utilize the multi-patch sampling scheme with 5 patches, following~\cite{few_shot_diff}.

For the downstream segmentation experiments, we utilized a publicly available U-Net implementation with a MiT-B2 transformer encoder~\cite{seg_pytorch,segformer}. We train a binary segmentation, with foreground vs background, which translates to tumor vs non-tumor for the histopathology datasets. In experiments with synthetic images, we combined real and synthetic images, oversampling synthetic images with a factor of four to one. As a loss function, we utilized a mix of cross-entropy loss and Dice loss, with weights 0.1 and 0.9, respectively. We trained for 75 epochs with 10,000 samples each and selected the model with the lowest validation loss for testing. Each experiment was performed five times to compute the mean and standard deviation.

\subsection{Evaluation}

For the evaluation, we first qualitatively assessed the generated images, focusing on color verification for the flower dataset. We aimed to verify whether the model produces images of the desired style and how it reacts to styles, especially colors, not seen during training. A similar qualitative assessment was performed for the histopathological dataset, although verifying correct styles is more challenging.

To assess the feature distribution of generated images, we employed the widely used \ac{fid} and \ac{is}. Both metrics utilize an Inception network as a feature extractor, which was trained on ImageNet. For the \ac{fid}, we sample real reference images from the style source data to validate whether our method creates images more similar to the data not seen during training of the diffusion model.

As final evaluation of the information content contained in the generated images, we report the results for the semi-supervised segmentation task. We focus on the \ac{iou} score of the foreground class as well as the variation of the \ac{iou} scores between the samples to demonstrate that the inclusion of synthetic data generated with our proposed approach improves generalization.

\subsection{Results and Discussions}

\label{sec:gen_results}

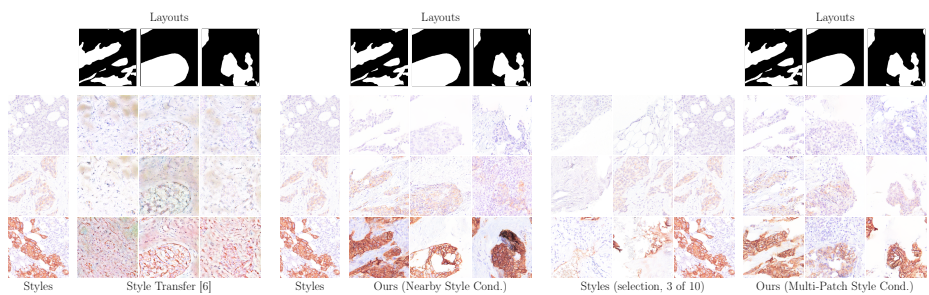
\begin{figure*}
    \centering
    \resizebox{1.0\textwidth}{!}{%
    \begin{tikzpicture}[ image/.style = {inner sep=0pt, outer sep=0pt}, node distance = 1mm and 1mm]
    \def\imageWidth{3cm}
    \def\maskWidth{2.8cm}
    
    \node [image] (frame1) {\includegraphics[width=\imageWidth]{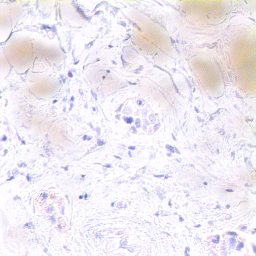}};
    \node [image,right=of frame1] (frame2) {\includegraphics[width=\imageWidth]{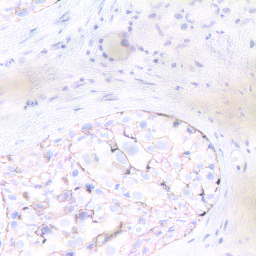}};
    \node [image,right=of frame2] (frame3) {\includegraphics[width=\imageWidth]{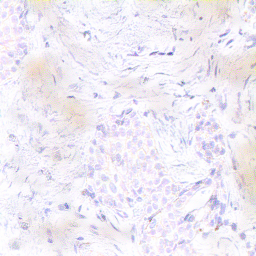}};

    \node[image,below=of frame1] (frame4) {\includegraphics[width=\imageWidth]{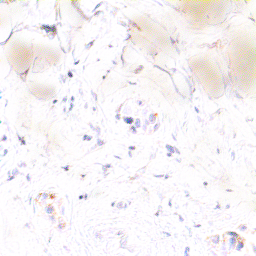}};
    \node[image,right=of frame4] (frame5) {\includegraphics[width=\imageWidth]{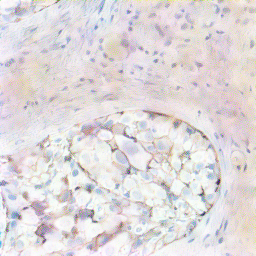}};
    \node[image,right=of frame5] (frame6) {\includegraphics[width=\imageWidth]{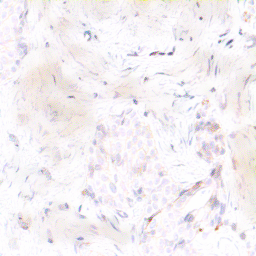}};

    \node[image,below=of frame4] (frame7) {\includegraphics[width=\imageWidth]{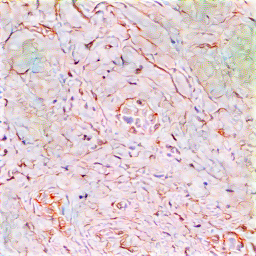}};
    \node[image,right=of frame7] (frame8) {\includegraphics[width=\imageWidth]{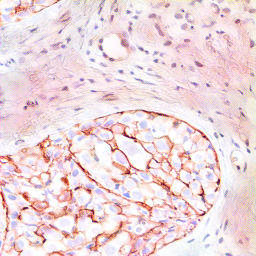}};
    \node[image,right=of frame8] (frame9) {\includegraphics[width=\imageWidth]{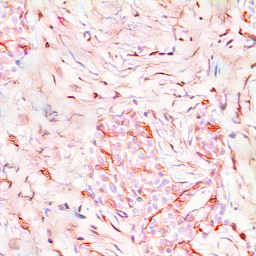}};

    \node[image,right=1.0cm of frame3] (nb_mask1) {\includegraphics[width=\imageWidth]{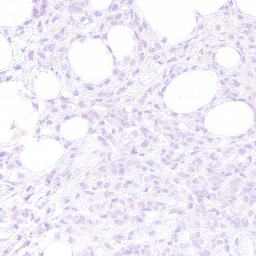}};
    \node[image,right=1.0cm of frame6] (nb_mask2) {\includegraphics[width=\imageWidth]{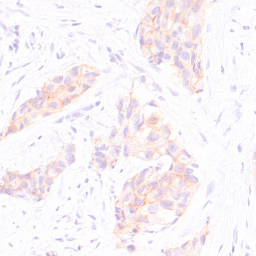}};
    \node[image,right=1.0cm of frame9] (nb_mask3) {\includegraphics[width=\imageWidth]{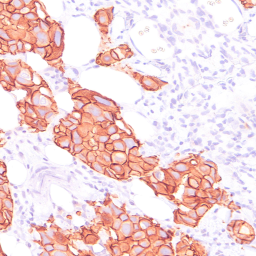}};

    \node [image,right=0.5cm of nb_mask1] (nearby1) {\includegraphics[width=\imageWidth]{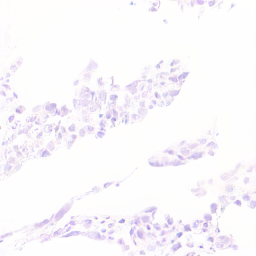}};
    \node [image,right=of nearby1] (nearby2) {\includegraphics[width=\imageWidth]{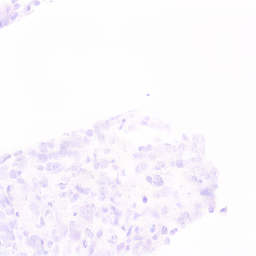}};
    \node [image,right=of nearby2] (nearby3) {\includegraphics[width=\imageWidth]{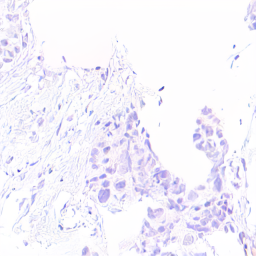}};

    \node[image,below=of nearby1] (nearby4) {\includegraphics[width=\imageWidth]{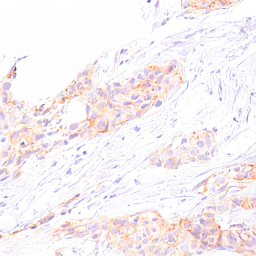}};
    \node[image,right=of nearby4] (nearby5) {\includegraphics[width=\imageWidth]{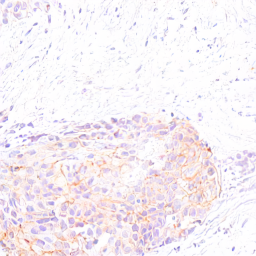}};
    \node[image,right=of nearby5] (nearby6) {\includegraphics[width=\imageWidth]{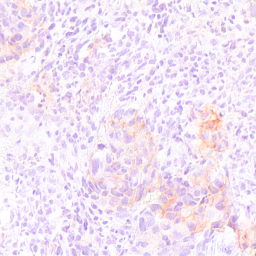}};

    \node[image,below=of nearby4] (nearby7) {\includegraphics[width=\imageWidth]{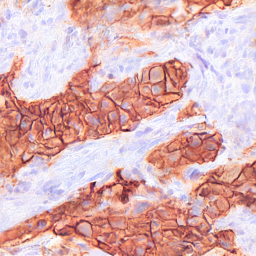}};
    \node[image,right=of nearby7] (nearby8) {\includegraphics[width=\imageWidth]{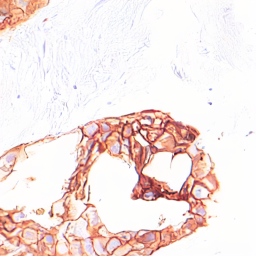}};
    \node[image,right=of nearby8] (nearby9) {\includegraphics[width=\imageWidth]{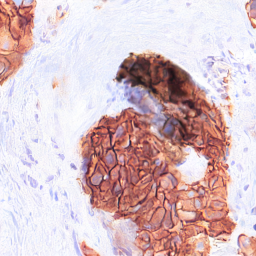}};

    \node[image,left=0.5cm of frame1] (mask1) {\includegraphics[width=\imageWidth]{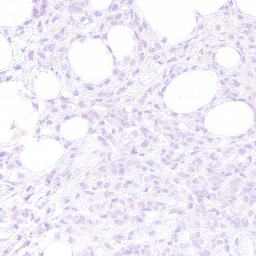}};
    \node[image,left=0.5cm of frame4] (mask2) {\includegraphics[width=\imageWidth]{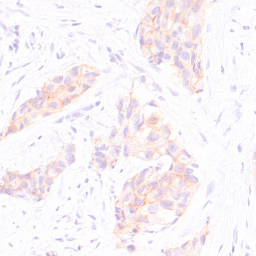}};
    \node[image,left=0.5cm of frame7] (mask3) {\includegraphics[width=\imageWidth]{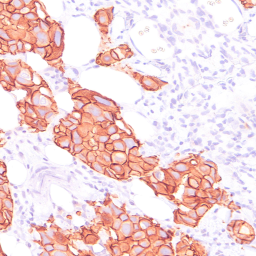}};

    \node[image,above=0.5cm of frame1,draw=black, line width=0.1cm] (style1) {\includegraphics[width=\maskWidth]{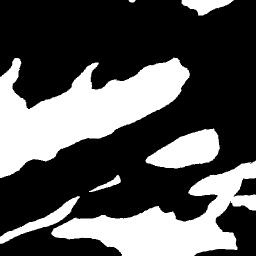}};
    \node[image,above=0.5cm of frame2,draw=black, line width=0.1cm] (style2) {\includegraphics[width=\maskWidth]{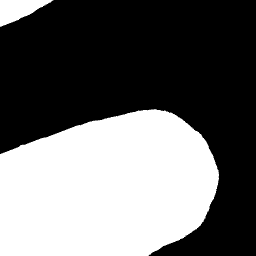}};
    \node[image,above=0.5cm of frame3,draw=black, line width=0.1cm] (style3) {\includegraphics[width=\maskWidth]{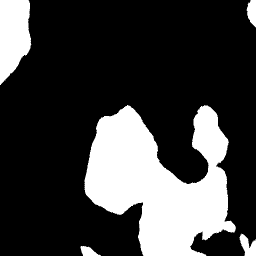}};

    \node[image,above=0.5cm of nearby1,draw=black, line width=0.1cm] (nb_style1) {\includegraphics[width=\maskWidth]{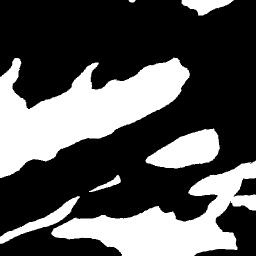}};
    \node[image,above=0.5cm of nearby2,draw=black, line width=0.1cm] (nb_style2) {\includegraphics[width=\maskWidth]{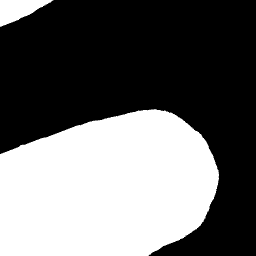}};
    \node[image,above=0.5cm of nearby3,draw=black, line width=0.1cm] (nb_style3) {\includegraphics[width=\maskWidth]{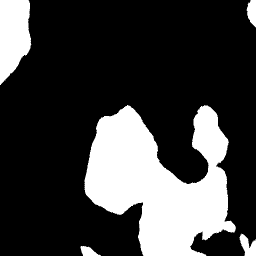}};

    \node[image,right=1cm of nearby3] (mp_style1) {\includegraphics[width=\imageWidth]{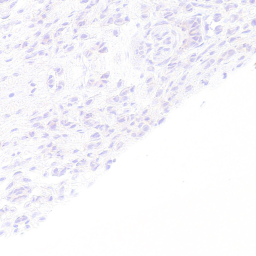}};
    \node[image,right=of mp_style1] (mp_style2) {\includegraphics[width=\imageWidth]{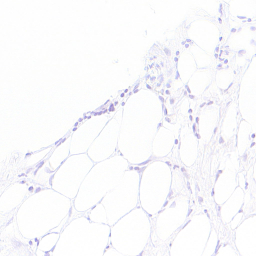}};
    \node[image,right=of mp_style2] (mp_style3) {\includegraphics[width=\imageWidth]{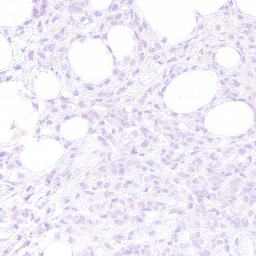}};

    \node[image,right=1cm of nearby6] (mp_style4) {\includegraphics[width=\imageWidth]{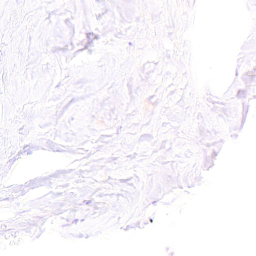}};
    \node[image,right=of mp_style4] (mp_style5) {\includegraphics[width=\imageWidth]{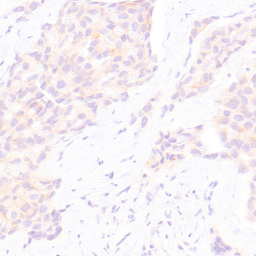}};
    \node[image,right=of mp_style5] (mp_style6) {\includegraphics[width=\imageWidth]{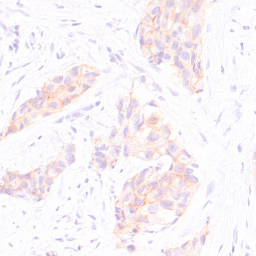}};

    \node[image,right=1cm of nearby9] (mp_style7) {\includegraphics[width=\imageWidth]{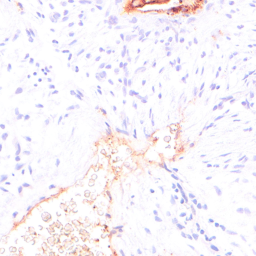}};
    \node[image,right=of mp_style7] (mp_style8) {\includegraphics[width=\imageWidth]{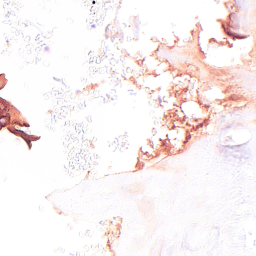}};
    \node[image,right=of mp_style8] (mp_style9) {\includegraphics[width=\imageWidth]{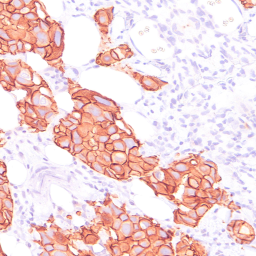}};

    \node [image,right=0.5cm of mp_style3] (mp1) {\includegraphics[width=\imageWidth]{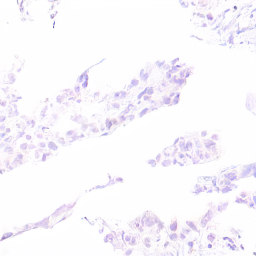}};
    \node [image,right=of mp1] (mp2) {\includegraphics[width=\imageWidth]{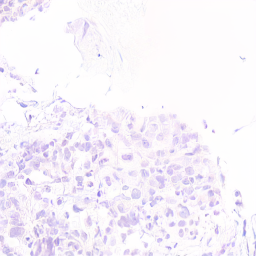}};
    \node [image,right=of mp2] (mp3) {\includegraphics[width=\imageWidth]{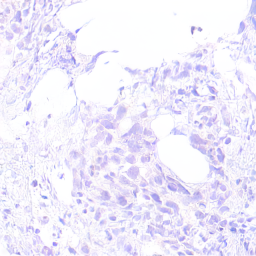}};

    \node[image,below=of mp1] (mp4) {\includegraphics[width=\imageWidth]{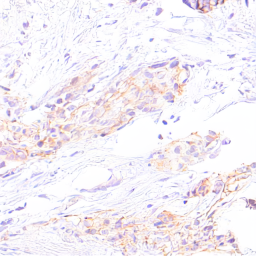}};
    \node[image,right=of mp4] (mp5) {\includegraphics[width=\imageWidth]{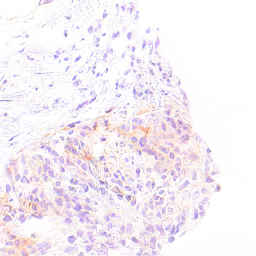}};
    \node[image,right=of mp5] (mp6) {\includegraphics[width=\imageWidth]{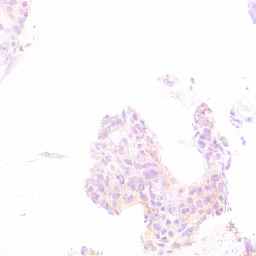}};

    \node[image,below=of mp4] (mp7) {\includegraphics[width=\imageWidth]{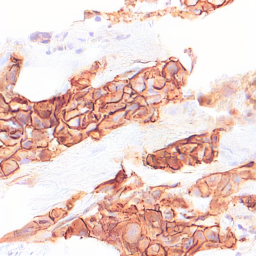}};
    \node[image,right=of mp7] (mp8) {\includegraphics[width=\imageWidth]{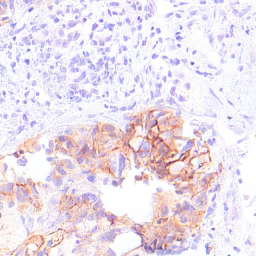}};
    \node[image,right=of mp8] (mp9) {\includegraphics[width=\imageWidth]{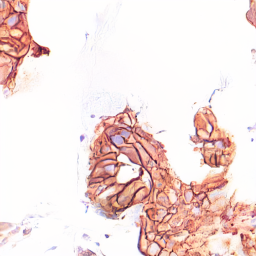}};

    \node[image,above=0.5cm of mp1, draw=black, line width=0.1cm] (mp_mask1) {\includegraphics[width=\maskWidth]{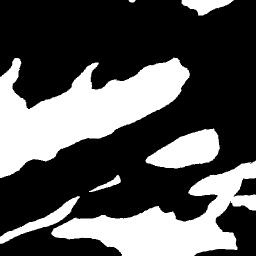}};
    \node[image,above=0.5cm of mp2, draw=black, line width=0.1cm] (mp_mask2) {\includegraphics[width=\maskWidth]{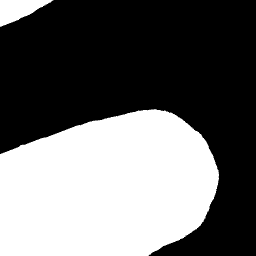}};
    \node[image,above=0.5cm of mp3, draw=black, line width=0.1cm] (mp_mask3) {\includegraphics[width=\maskWidth]{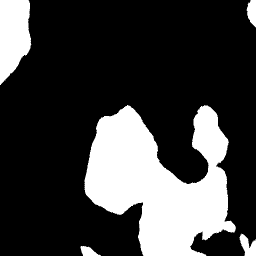}};

    \node[yshift = -2.0cm, inner sep=0] (text1) at ($(frame7)!0.5!(frame9)$) {\LARGE{Style Transfer~\cite{baseline_style}}};
    \node[yshift = -2.0cm, inner sep=0] (text1) at ($(nearby7)!0.5!(nearby9)$) {\LARGE{Ours (Nearby Style Cond.)}};
    \node[yshift = -2.0cm, inner sep=0] (text1) at ($(mp7)!0.5!(mp9)$) {\LARGE{Ours (Multi-Patch Style Cond.)}};
    \node[inner sep=0] (text2) at (mask3|- text1) {\LARGE{Styles}};
    \node[inner sep=0] (text2) at (nb_mask3|- text1) {\LARGE{Styles}};
    \node[inner sep=0] (text2) at (mp_style8|- text1) {\LARGE{Styles (selection, 3 of 10)}};
    \node[yshift = 2.0cm] (text3) at ($(style1)!0.5!(style3)$) {\LARGE{Layouts}};
    \node[yshift = 2.0cm] (text3) at ($(nb_style1)!0.5!(nb_style3)$) {\LARGE{Layouts}};
    \node[yshift = 2.0cm] (text3) at ($(mp_mask1)!0.5!(mp_mask3)$) {\LARGE{Layouts}};
    \end{tikzpicture}}
    \caption{Image generation results with the HER2 dataset, for the style transfer baseline~\cite{baseline_style} (left), our proposed method trained with nearby patches as style source (center) and our proposed method trained with multi-patches as style source (right). Note that white represents tumor tissue in the layout images, while black includes all background structures.}
    \label{fig:her2_gen}
\end{figure*}

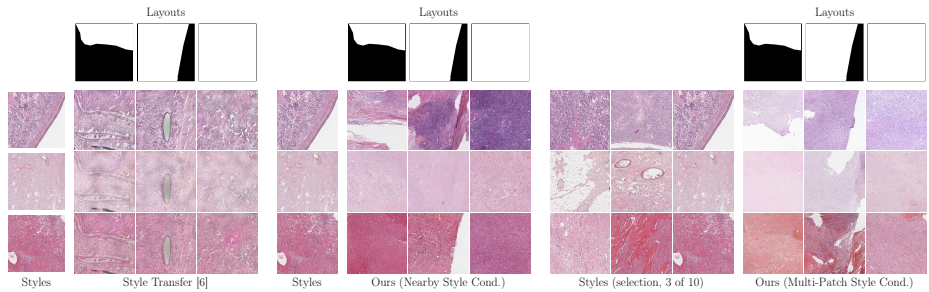
\begin{figure*}[t]
    \centering
    \resizebox{1.0\textwidth}{!}{%
    \begin{tikzpicture}[ image/.style = {inner sep=0pt, outer sep=0pt}, node distance = 1mm and 1mm]
    \def\imageWidth{3cm}
    \def\maskWidth{2.8cm}
    
    \node [image] (frame1) {\includegraphics[width=\imageWidth]{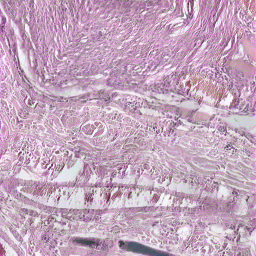}};
    \node [image,right=of frame1] (frame2) {\includegraphics[width=\imageWidth]{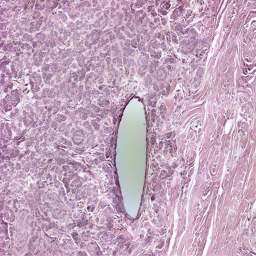}};
    \node [image,right=of frame2] (frame3) {\includegraphics[width=\imageWidth]{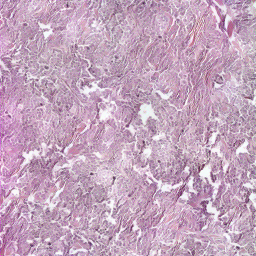}};

    \node[image,below=of frame1] (frame4) {\includegraphics[width=\imageWidth]{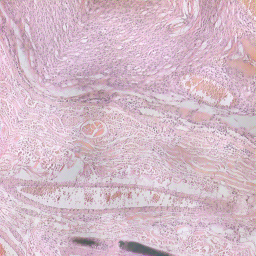}};
    \node[image,right=of frame4] (frame5) {\includegraphics[width=\imageWidth]{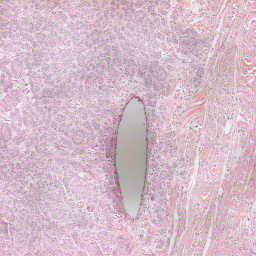}};
    \node[image,right=of frame5] (frame6) {\includegraphics[width=\imageWidth]{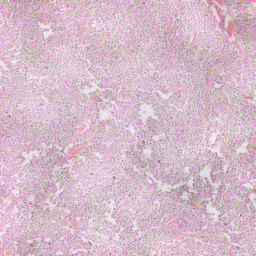}};

    \node[image,below=of frame4] (frame7) {\includegraphics[width=\imageWidth]{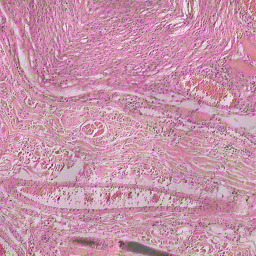}};
    \node[image,right=of frame7] (frame8) {\includegraphics[width=\imageWidth]{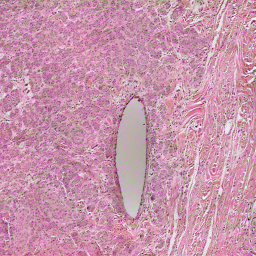}};
    \node[image,right=of frame8] (frame9) {\includegraphics[width=\imageWidth]{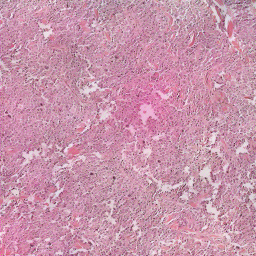}};

    \node[image,right=1.0cm of frame3] (nb_mask1) {\includegraphics[width=\imageWidth]{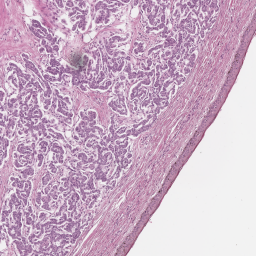}};
    \node[image,right=1.0cm of frame6] (nb_mask2) {\includegraphics[width=\imageWidth]{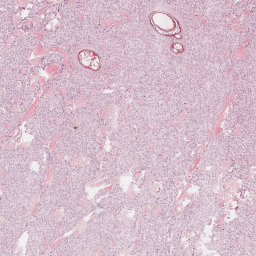}};
    \node[image,right=1.0cm of frame9] (nb_mask3) {\includegraphics[width=\imageWidth]{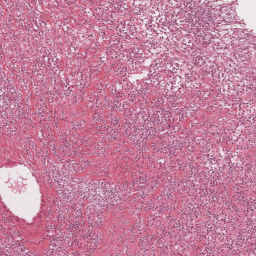}};

    \node [image,right=0.5cm of nb_mask1] (nearby1) {\includegraphics[width=\imageWidth]{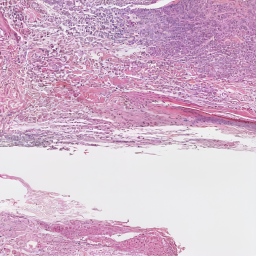}};
    \node [image,right=of nearby1] (nearby2) {\includegraphics[width=\imageWidth]{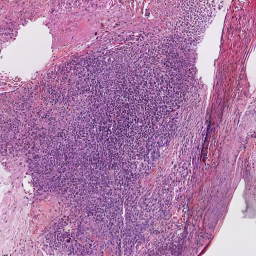}};
    \node [image,right=of nearby2] (nearby3) {\includegraphics[width=\imageWidth]{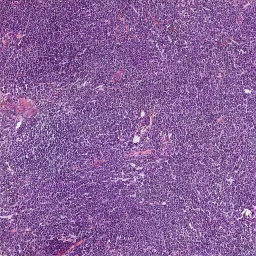}};

    \node[image,below=of nearby1] (nearby4) {\includegraphics[width=\imageWidth]{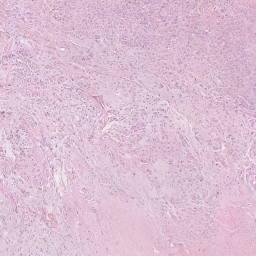}};
    \node[image,right=of nearby4] (nearby5) {\includegraphics[width=\imageWidth]{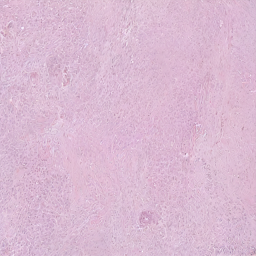}};
    \node[image,right=of nearby5] (nearby6) {\includegraphics[width=\imageWidth]{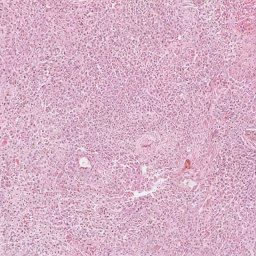}};

    \node[image,below=of nearby4] (nearby7) {\includegraphics[width=\imageWidth]{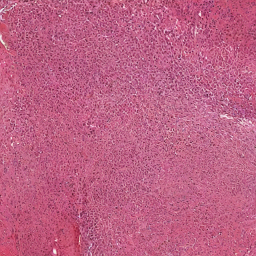}};
    \node[image,right=of nearby7] (nearby8) {\includegraphics[width=\imageWidth]{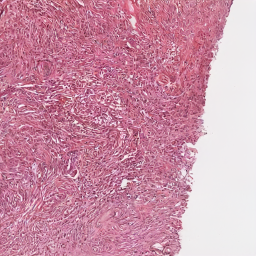}};
    \node[image,right=of nearby8] (nearby9) {\includegraphics[width=\imageWidth]{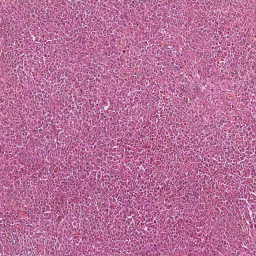}};

    \node[image,left=0.5cm of frame1] (mask1) {\includegraphics[width=\maskWidth]{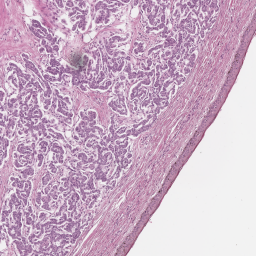}};
    \node[image,left=0.5cm of frame4] (mask2) {\includegraphics[width=\maskWidth]{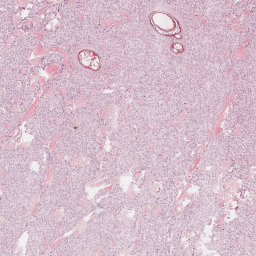}};
    \node[image,left=0.5cm of frame7] (mask3) {\includegraphics[width=\maskWidth]{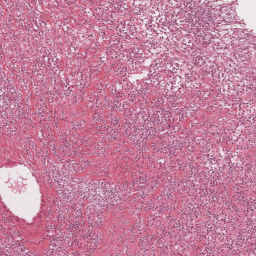}};

    \node[image,above=0.5cm of frame1,draw=black, line width=0.1cm] (style1) {\includegraphics[width=\maskWidth]{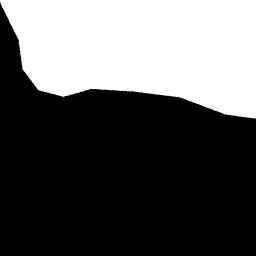}};
    \node[image,above=0.5cm of frame2,draw=black, line width=0.1cm] (style2) {\includegraphics[width=\maskWidth]{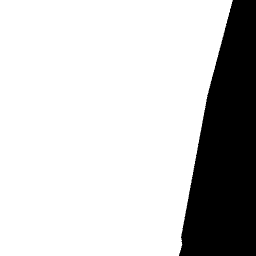}};
    \node[image,above=0.5cm of frame3,draw=black, line width=0.1cm] (style3) {\includegraphics[width=\maskWidth]{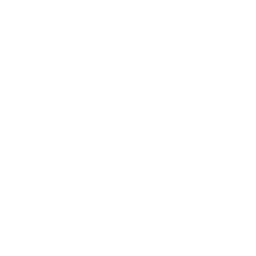}};

    \node[image,above=0.5cm of nearby1,draw=black, line width=0.1cm] (nb_style1) {\includegraphics[width=\maskWidth]{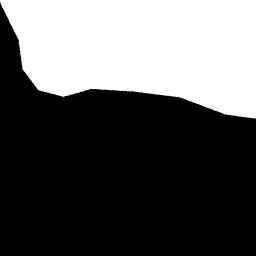}};
    \node[image,above=0.5cm of nearby2,draw=black, line width=0.1cm] (nb_style2) {\includegraphics[width=\maskWidth]{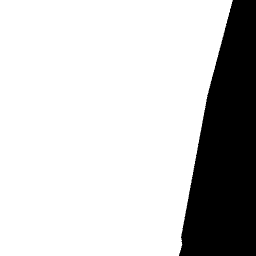}};
    \node[image,above=0.5cm of nearby3,draw=black, line width=0.1cm] (nb_style3) {\includegraphics[width=\maskWidth]{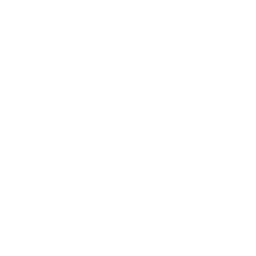}};

    \node[image,right=1cm of nearby3] (mp_style1) {\includegraphics[width=\imageWidth]{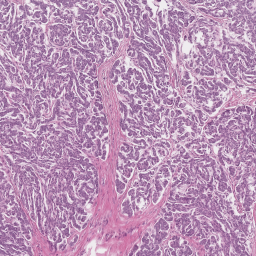}};
    \node[image,right=of mp_style1] (mp_style2) {\includegraphics[width=\imageWidth]{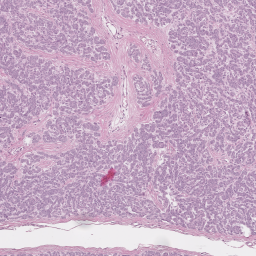}};
    \node[image,right=of mp_style2] (mp_style3) {\includegraphics[width=\imageWidth]{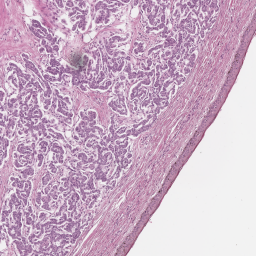}};

    \node[image,right=1cm of nearby6] (mp_style4) {\includegraphics[width=\imageWidth]{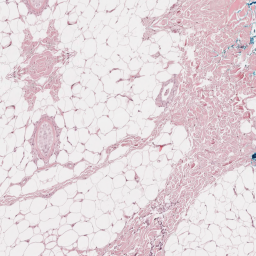}};
    \node[image,right=of mp_style4] (mp_style5) {\includegraphics[width=\imageWidth]{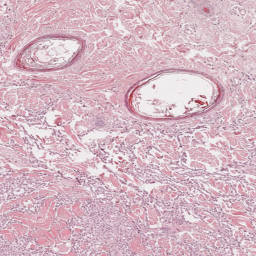}};
    \node[image,right=of mp_style5] (mp_style6) {\includegraphics[width=\imageWidth]{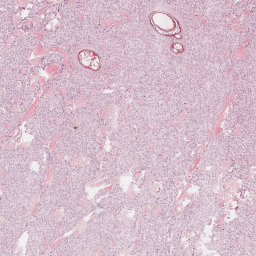}};

    \node[image,right=1cm of nearby9] (mp_style7) {\includegraphics[width=\imageWidth]{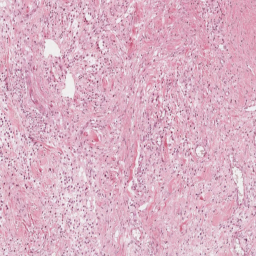}};
    \node[image,right=of mp_style7] (mp_style8) {\includegraphics[width=\imageWidth]{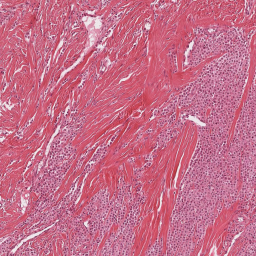}};
    \node[image,right=of mp_style8] (mp_style9) {\includegraphics[width=\imageWidth]{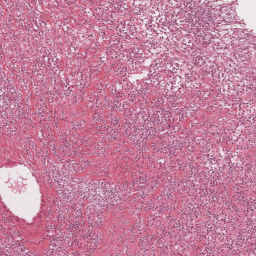}};

    \node [image,right=0.5cm of mp_style3] (mp1) {\includegraphics[width=\imageWidth]{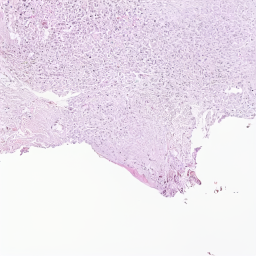}};
    \node [image,right=of mp1] (mp2) {\includegraphics[width=\imageWidth]{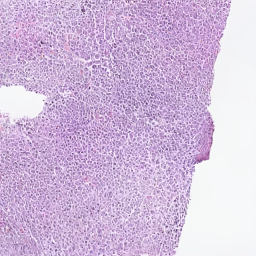}};
    \node [image,right=of mp2] (mp3) {\includegraphics[width=\imageWidth]{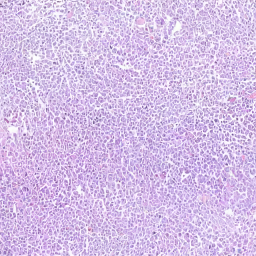}};

    \node[image,below=of mp1] (mp4) {\includegraphics[width=\imageWidth]{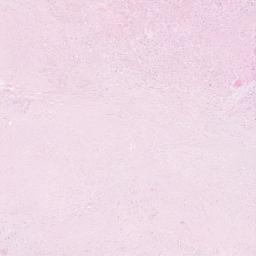}};
    \node[image,right=of mp4] (mp5) {\includegraphics[width=\imageWidth]{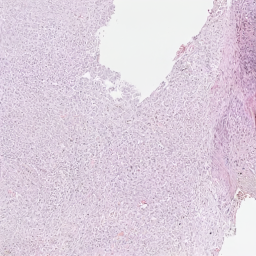}};
    \node[image,right=of mp5] (mp6) {\includegraphics[width=\imageWidth]{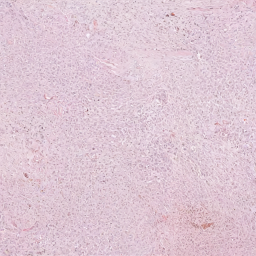}};

    \node[image,below=of mp4] (mp7) {\includegraphics[width=\imageWidth]{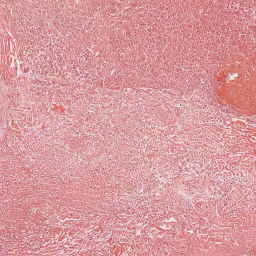}};
    \node[image,right=of mp7] (mp8) {\includegraphics[width=\imageWidth]{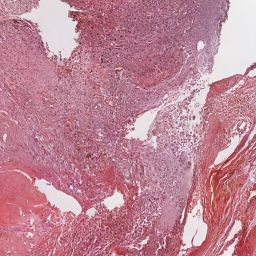}};
    \node[image,right=of mp8] (mp9) {\includegraphics[width=\imageWidth]{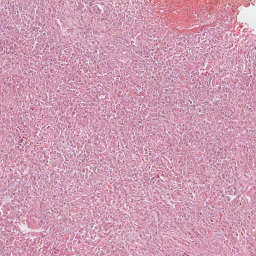}};

    \node[image,above=0.5cm of mp1, draw=black, line width=0.1cm] (mp_mask1) {\includegraphics[width=\maskWidth]{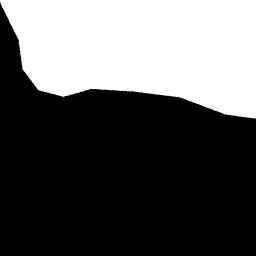}};
    \node[image,above=0.5cm of mp2, draw=black, line width=0.1cm] (mp_mask2) {\includegraphics[width=\maskWidth]{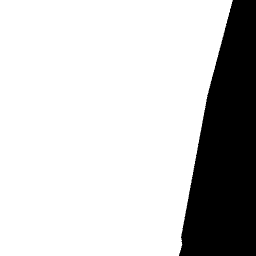}};
    \node[image,above=0.5cm of mp3, draw=black, line width=0.1cm] (mp_mask3) {\includegraphics[width=\maskWidth]{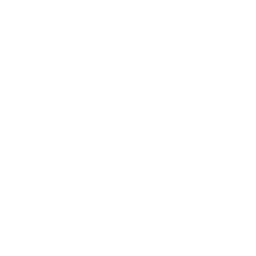}};

    \node[yshift = -2.0cm, inner sep=0] (text1) at ($(frame7)!0.5!(frame9)$) {\LARGE{Style Transfer~\cite{baseline_style}}};
    \node[yshift = -2.0cm, inner sep=0] (text1) at ($(nearby7)!0.5!(nearby9)$) {\LARGE{Ours (Nearby Style Cond.)}};
    \node[yshift = -2.0cm, inner sep=0] (text1) at ($(mp7)!0.5!(mp9)$) {\LARGE{Ours (Multi-Patch Style Cond.)}};
    \node[inner sep=0] (text2) at (mask3|- text1) {\LARGE{Styles}};
    \node[inner sep=0] (text2) at (nb_mask3|- text1) {\LARGE{Styles}};
    \node[inner sep=0] (text2) at (mp_style8|- text1) {\LARGE{Styles (selection, 3 of 10)}};
    \node[yshift = 2.0cm] (text3) at ($(style1)!0.5!(style3)$) {\LARGE{Layouts}};
    \node[yshift = 2.0cm] (text3) at ($(nb_style1)!0.5!(nb_style3)$) {\LARGE{Layouts}};
    \node[yshift = 2.0cm] (text3) at ($(mp_mask1)!0.5!(mp_mask3)$) {\LARGE{Layouts}};
    \end{tikzpicture}}
    \caption{Image generation results with the CATCH dataset, for the style transfer baseline~\cite{baseline_style} (left), our proposed method trained with nearby patches as style source (center) and our proposed method trained with multi-patches as style source (right). Note that white represents tumor tissue in the layout images, while black includes all background structures.}
    \label{fig:catch_gen}
\end{figure*}

\noindent\textbf{Qualitative image generation results.}
\Cref{fig:flowers_gen} shows synthetic images for the flower dataset. For the style transfer~\cite{baseline_style}, the color of the requested style image is present, but it is not located correctly with respect to the flower. With the semantic conditioned diffusion model, the colors of the flowers were chosen by the network and are representative of the color distribution in the training data, where flowers with the colors blue, purple, and pink are underrepresented or excluded. Our method generates images that adhere to the requested layout, and flowers with colors similar to the ones in the style query images. This is noteworthy since these colors were excluded from training, and indicates that the model was able to pick up color as a style concept.

\Cref{fig:her2_gen,fig:catch_gen} show examples of generated images for the \ac{her2} and \ac{catch} dataset. The style transfer baseline~\cite{baseline_style} captures some of the staining characteristics for the \ac{her2} dataset, but also creates some darker blurred areas, which are not commonly seen in \ac{her2} images. On the \ac{catch} dataset, the global color scheme is transferred to the synthetic images, but no delineation between tumor and background area is visible. For our method the generated images adhere to the spatial layout and also represent the styles in the query images for both the nearby and multi-patch configuration, even though the query images were not seen during training.

\begin{table}[t]\setlength{\tabcolsep}{8pt}
\centering
\resizebox{0.75\columnwidth}{!}{%
\begin{tabular}{ccccccccc}
\toprule
& Method & \multicolumn{3}{c}{FID$\downarrow$} && \multicolumn{3}{c}{IS$\uparrow$} \\
\cmidrule(l){3-5}\cmidrule(l){6-9}
&
&
\multicolumn{7}{c}{960 / 480 / 144 Train Images} \\
\multirow{4}{*}{\rotatebox[origin=c]{90}{\textbf{Flowers}}} & Style Transfer     & 105.2 & -- & -- && 5.84 & -- & -- \\
& Semantic DM        & 81.66 & 80.78 & 78.77 && 4.36 & 3.89 & 4.28 \\
& Nearby (ours)      & 76.04 & 56.78 & 63.19 && 3.50 & 3.46 & 3.19 \\
& Multi-patch (ours) & 86.60 & 57.62 & 77.68 && 3.47 & 3.48 & 3.32 \\
\cmidrule(l){3-5}\cmidrule(l){6-9}
&
&
\multicolumn{7}{c}{24 / 12 / 6 Train WSI} \\
\multirow{4}{*}{\rotatebox[origin=c]{90}{\textbf{HER2}}} &
Style Transfer     & 46.45 & -- & -- && 3.63 & -- & -- \\
& Swapping AE        & 82.76 & 72.27 & 77.49 && 2.64 & 2.71 & 2.66 \\
& Semantic DM        & 46.50 & 46.43 & 47.01 && 3.58 & 3.57 & 3.54 \\
& sViT        & 56.68 & 50.42 & 48.69 && 3.04 & 3.19 & 3.37 \\
& Nearby (ours)      & 45.75 & 44.41 & 46.14 && 3.60 & 3.46 & 3.54 \\
& Multi-patch (ours) & 45.22 & 45.58 & 46.47 && 3.59 & 3.39 & 3.55 \\
\cmidrule(l){3-5}\cmidrule(l){6-9}
&
&
\multicolumn{7}{c}{42 / 21 / 7 Train WSI} \\
\multirow{4}{*}{\rotatebox[origin=c]{90}{\textbf{CATCH}}} 
& Style Transfer     & 89.70 & -- & -- && 3.65 & -- & -- \\
& Swapping AE        & 89.97 & 93.39 & 101.48 && 3.64 & 3.65 & 3.38 \\
& Semantic DM        & 85.06 & 89.69 & 113.16 && 3.51 & 3.31 & 2.95 \\
& sViT               & 86.02 & 89.49 & 108.35 && 3.57 & 3.64 & 3.66 \\
& Nearby (ours)      & 84.09 & 81.83 & 102.98 && 3.50 & 3.42 & 3.24 \\
& Multi-patch (ours) & 84.24 & 89.61 & 106.60 && 3.52 & 3.32 & 3.07 \\
\bottomrule
\end{tabular}
}
\caption{\ac{fid} and \ac{is} scores for the three datasets, different generation methods, and different amounts of training data.}
\label{table:img_gen}
\end{table}

\noindent\textbf{Quality metrics.}
\Cref{table:img_gen} shows the \ac{fid} and \ac{is} metrics for the generated images. Style transfer scores are only reported for one setting, as the method is not trained and produces the same results independent of the amount of training data. SwappingAE and sViT experiments were performed for the histopathology datasets. For all datasets and methods, no clear trend is visible for the \ac{is}. The \ac{fid} scores of our proposed methods are, for the most cases, lower than those of the baseline methods, although only relatively minor difference are present in general, with the exception of Swapping AE for the HER2 dataset. These results hint at the capability of our method to create images that are adapted to the data distribution of style source images, which served as reference for the \ac{fid} score.

\noindent\textbf{Discussion of generation results.}
For the flower dataset, the color is the most dominant feature. Although we expect the diffusion network to have learned some bias towards the shape of the flowers, we see no major influence of that, which we expect to be caused by the effect of the classifier-free guidance.

\begin{table*}[ht]
\resizebox{\textwidth}{!}{%
\begin{tabular}{clccccccccccc}
\toprule
 & Synthetic data  & Mean IoU && IoU Variance && Mean IoU && IoU Variance && Mean IoU && IoU Variance \\
\cmidrule{3-3}\cmidrule{5-5}\cmidrule{7-7}\cmidrule{9-9}\cmidrule{11-11}\cmidrule{13-13}
&& \multicolumn{3}{c}{24 WSI} && \multicolumn{3}{c}{12 WSI} && \multicolumn{3}{c}{4 WSI} \\
\cmidrule{3-5}\cmidrule{7-9}\cmidrule{11-13}
\multicolumn{1}{c}{\multirow{5}{*}{\rotatebox[origin=c]{90}{\textbf{HER2}}}} &  None               & 78.88 (0.63)                 && 0.33 (0.08)                      && 74.63 (1.11)                && 1.57 (0.64)                      && 74.17 (0.54)                && 1.84 (0.44)                       \\
& Style Transfer     & 78.67 (0.11)                 && 0.45 (0.04)      &&    75.78 (0.75)           &&   1.07 (0.30)          &&     74.60 (1.08)        &&       2.01 (0.71)                            \\
& Swapping AE     & 77.06 (0.99)              &&     0.43 (0.05)       &&    76.00 (0.69)        &&   0.53 (0.09)   &&  74.72 (0.80)   &&   0.71 (0.49)       \\
& Semantic DM        & 79.15 (0.86)                 && 0.32 (0.03)                      && 77.28 (0.81)               && 0.39 (0.11)                       && 76.77 (0.19)                 && 0.66 (0.14)                       \\
& sViT     & 78.88 (0.92)                 && 0.41 (0.17)        &&  75.45 (0.27)     &&     1.01 (0.20) && 76.55 (0.81)   &&     0.86 (0.26)  \\
& Nearby (ours)      & \textbf{79.78 (0.15)}        && \textbf{0.26 (0.01)}              && 77.72 (1.05)                 && 0.36 (0.07)                       && 77.61 (0.50)                 && 0.28 (0.08)                       \\
& Multi-patch (ours) & 79.54 (0.82)                 && 0.26 (0.03)                       && \textbf{78.43 (0.63)}        && \textbf{0.33 (0.11)}              && \textbf{78.81 (0.25)}        && \textbf{0.24 (0.03)}              \\ 
\cmidrule{3-13}
&& \multicolumn{3}{c}{42 WSI} && \multicolumn{3}{c}{21 WSI} && \multicolumn{3}{c}{7 WSI} \\
\cmidrule{3-5}\cmidrule{7-9}\cmidrule{11-13}

 \multicolumn{1}{c}{\multirow{5}{*}{\rotatebox[origin=c]{90}{\textbf{CATCH}}}}  & None               & 88.07 (0.21)                 && 2.42 (0.22)                     && 87.66 (0.35)               && 3.54 (0.08)                      && 85.71 (0.35)                && 3.43 (0.10)                       \\
& Style Transfer     & 87.47 (0.23)                 && 2.00 (0.10)                       &&     86.87 (0.59)     &&     3.75 (0.26)                              &&  84.68 (0.55)   &&     3.46 (0.21)                              \\
& Swapping AE     & 87.53 (0.65)                 && 3.06 (0.24)                      &&   85.50 (1.01)    &&    4.89 (0.39)      &&   83.89 (0.85)    &&    3.77 (0.25)             \\
& Semantic DM        & 87.72 (0.89)                 && 2.68 (0.15)                       && 86.85 (0.28)                 && 3.63 (0.10)                       && 85.62 (0.50)                 && 3.48 (0.12)                       \\
& sViT     & 87.97 (0.71)                && 2.56 (0.13)                      &&   86.84 (0.50)    &&    3.52 (0.25)          &&   85.61 (1.51)        &&  3.57 (0.26)   \\
& Nearby (ours)      & 88.02 (0.38)                 && 2.41 (0.11)                       && 87.46 (0.55)                 && 3.45 (0.21)                       && \textbf{86.61 (0.80)}        && \textbf{3.18 (0.17)}              \\
& Multi-patch (ours) & \textbf{88.09 (0.27)}        && \textbf{2.36 (0.06)}              && \textbf{87.72 (0.31)}        && \textbf{3.12 (0.04)}              && 85.09 (0.69)                 && 3.55 (0.15)                       \\ 

\bottomrule
\end{tabular}
}
\caption{Segmentation results for the histopathological datasets, with different amounts of training data and synthetic images.}
\label{table:seg_results}
\end{table*}

Histopathological images generated with our method with nearby style conditioning show a strong resemblance to the unseen style queries. While this works for cases where all relevant information is contained in the style query image, we expect our method to recreate styles from the training distribution if style information is missing in the query images, \emph{e.g.}, if the layout requests the presence of tumor tissue, but no tumor is included in the style query image. In the supplementary material we show examples of generated images where style information is missing. We see this issue reduced for multi-patches style conditioning, since the network appears to be able to compose the style information from multiple input patches. 
Although the quality metrics show promising results for our generated images, we emphasize that these results should be interpreted carefully and follow-up studies are needed~\cite{barratt2018note,BORJI2022103329}. For example, the feature extraction of the underlying InceptionNet could be more sensitive to the layout than to the style. Similarly, we measure the \ac{fid} score compared to unseen data which is from the same site as the training data and is expected to have substantial similarity. Therefore, a low \ac{fid} could reflect good quality images rather than attaining specific style nuances of the style source data.

\noindent\textbf{Semi-supervised segmentation results.}
The segmentation results in \cref{table:seg_results} indicate no or only a minimal benefit from using additional images generated with the baseline style transfer method~\cite{baseline_style} for training. Adding images generated with the Swapping Autoencoder\cite{style_swapping} provided minor benetifs only for the lower data settings of the \ac{her2} dataset. Conversely, adding synthetic images from a semantic conditioned diffusion model improved mean IoU results only for the \ac{her2} dataset. Utilizing the sViT as style encoder performed inferior to our method in all cases.
The introduction of synthetic data from our proposed method demonstrated benefits for all cases, with particularly positive outcomes observed for multi-patch style conditioning. Our method effectively closes 98\% of the performance gap in mean IoU between using 24 \acp{wsi} and using 6 \acp{wsi} for training with the \ac{her2} dataset. In a similar setup with the \ac{catch} dataset, we reduce the performance gap by 38\%. Regarding IoU variation between samples, notable improvements were observed, especially for the \ac{her2} dataset at lower data settings, indicating a more balanced segmentation performance.

Semi-supervised segmentation experiments were also conducted for the flower dataset, and the results are reported in the supplementary material. No improvements were observed with any synthetic data, as the segmentation results remained consistently high, even when the majority of data was removed.

\noindent\textbf{Discussion on semi-supervised segmentation results.} For the \ac{her2} dataset we saw benefits in both mean IoU and IoU variance between patients. Since there are only eight patients in the test set, we expect that the ``style'' of a subset of these patients is not well captured by the annotated data, but can be captured better with the synthetic images generated with our method leveraging the style source images from the additional unannotated data.
For the \ac{catch} dataset our method improved the segmentation compared to the baseline, though less prominently than for the \ac{her2} dataset. We argue this to be caused by more subtle style differences between training and test data, as well as a generally more robust segmentation task, offering less room for improvement.
While we showed in \cref{sec:gen_results} that our method is able to generate images with unseen styles, we do not see a benefit in segmentation performance for the flower dataset. We argue this to be caused by the relatively simple nature of the segmentation task, where even few images are sufficient to finetune our (ImageNet-pretrained) segmentation architecture.

\noindent\textbf{Limitations.} 
One limitation of our work regarding the flower dataset is a correlation between the layout and the style of images. In cases like this, the network may not fully separate layout and styles, therefore potentially ignoring the style information. We argue that classifier-free guidance reduces the impact of a shape bias, but further experiments are required to validate this assumption.
We also note that we considered style mappings between subtypes of the same shared category, \emph{e.g.}, different kinds of flowers, or different tumor subtypes for the same staining and origin. When deviating too far from the underlying style concepts, we do not expect our model to be able to create images with reasonable styles.
\section{Conclusion}
\label{sec:conclusion}

In this work, we propose \acf{stedm}, a novel method to create images with a known target variable but new unseen styles. The method includes a style encoder which extracts style information from style images. We also introduced an aggregation block, which allows the style information to be assembled from multiple style image.
Our experiments confirm the ability of our model to create images that resemble the requested styles and that carry fine-grained style information from the unseen style images. We quantify this with a semi-supervised histopathology segmentation task, which shows that our method is a potent way to make use unannotated data. Future work will include more extensive quantification of image quality. Test-time adaption may be an interesting further avenue to explore with this form of style conditioning.

\section{Acknowledgement}
This project is supported by the Bavarian State Ministry of Health and Care, project grants No. PBN-MGP-2010-0004-DigiOnko and PBN-MGP-2008-0003-DigiOnko. We also gratefully acknowledge the support from the Interdisciplinary Center for Clinical Research (IZKF, Clinician Scientist Program) of the Medical Faculty FAU Erlangen-Nürnberg. The authors gratefully acknowledge the scientific support and HPC resources provided by the Erlangen National High Performance Computing Center (NHR@FAU) of the Friedrich-Alexander-Universität Erlangen-Nürnberg (FAU) under the NHR project b160dc. NHR funding is provided by federal and Bavarian state authorities. NHR@FAU hardware is partially funded by the German Research Foundation (DFG) – 440719683. K.B., F.W. and M.Ö. acknowledge support by the German Research Foundation (DFG) project 460333672 CRC1540EBM. K.B. further acknowledges support by d.hip campus - Bavarian aim in form of a faculty endowment.

%
%
\bibliographystyle{splncs04}
\bibliography{main}

\begin{acronym}[type=hidden] 
\acro{her2}[HER2]{Human Epidermal growth factor Receptor 2} 
\acro{wsi}[WSI]{Whole Slide Image}
\acro{gan}[GAN]{Generative Adversarial Networks}
\acro{ldm}[LDM]{Latent Diffusion Model}
\acro{stedm}[STEDM]{Style-Extracting Diffusion Models}
\acro{ddpm}[DDPM]{Denoising Diffusion Probabilistic Model}
\acro{her2}[HER2]{Human Epidermal growth factor Receptor 2}
\acro{roi}[ROI]{Region of Interest}
\acrodefplural{roi}[ROIs]{Regions of Interest}
\acro{catch}[CATCH]{Pan-tumor CAnine CuTaneous Cancer Histology dataset}
\acro{fid}[FID]{Fréchet Inception Distance}
\acro{is}[IS]{Inception Score}
\acro{iou}[IoU]{Intersection over Union}
\end{acronym}

\clearpage
\appendix

\section{Flower dataset split}
\label{sec:flower_split}

\begin{figure*}
    \centering
    \resizebox{1.0\textwidth}{!}{%
    \begin{tikzpicture}[ image/.style = {inner sep=0pt, outer sep=0pt}, node distance = 1mm and 1mm]
    \def\imageWidth{1cm}

    \def\annoVals{4,5,6,7,10,11,14,15,19,20,22,23,25,26,30,34,37,41,42,43,45,46,47,48,49,50,52,53,55,57,58,59,60,62,63,64,68,69,70,73,74,80,88,89,98,99,100,101}

    \node[draw, align=center, minimum height = \imageWidth, minimum width = \imageWidth] (anno4){\Large{4}};
    \node[draw,below=0.2cm of anno4, align=center, minimum height = \imageWidth, minimum width = \imageWidth] (anno5){\Large{5}};
    \node[draw,below=0.2cm of anno5, align=center, minimum height = \imageWidth, minimum width = \imageWidth] (anno6){\Large{6}};
    \node[draw,below=0.2cm of anno6, align=center, minimum height = \imageWidth, minimum width = \imageWidth] (anno7){\Large{7}};
    \node[draw,below=0.2cm of anno7, align=center, minimum height = \imageWidth, minimum width = \imageWidth] (anno10){\Large{10}};
    \node[draw,below=0.2cm of anno10, align=center, minimum height = \imageWidth, minimum width = \imageWidth] (anno11){\Large{11}};
    \node[draw,below=0.2cm of anno11, align=center, minimum height = \imageWidth, minimum width = \imageWidth] (anno14){\Large{14}};
    \node[draw,below=0.2cm of anno14, align=center, minimum height = \imageWidth, minimum width = \imageWidth] (anno15){\Large{15}};
    
    \node[draw,right=3.5cm of anno4, align=center, minimum height = \imageWidth, minimum width = \imageWidth] (anno19){\Large{19}};
    \node[draw,below=0.2cm of anno19, align=center, minimum height = \imageWidth, minimum width = \imageWidth] (anno20){\Large{20}};
    \node[draw,below=0.2cm of anno20, align=center, minimum height = \imageWidth, minimum width = \imageWidth] (anno22){\Large{22}};
    \node[draw,below=0.2cm of anno22, align=center, minimum height = \imageWidth, minimum width = \imageWidth] (anno23){\Large{23}};
    \node[draw,below=0.2cm of anno23, align=center, minimum height = \imageWidth, minimum width = \imageWidth] (anno25){\Large{25}};
    \node[draw,below=0.2cm of anno25, align=center, minimum height = \imageWidth, minimum width = \imageWidth] (anno26){\Large{26}};
    \node[draw,below=0.2cm of anno26, align=center, minimum height = \imageWidth, minimum width = \imageWidth] (anno30){\Large{30}};
    \node[draw,below=0.2cm of anno30, align=center, minimum height = \imageWidth, minimum width = \imageWidth] (anno34){\Large{34}};
    
    \node[draw,right=3.5cm of anno19, align=center, minimum height = \imageWidth, minimum width = \imageWidth] (anno37){\Large{37}};
    \node[draw,below=0.2cm of anno37, align=center, minimum height = \imageWidth, minimum width = \imageWidth] (anno41){\Large{41}};
    \node[draw,below=0.2cm of anno41, align=center, minimum height = \imageWidth, minimum width = \imageWidth] (anno42){\Large{42}};
    \node[draw,below=0.2cm of anno42, align=center, minimum height = \imageWidth, minimum width = \imageWidth] (anno43){\Large{43}};
    \node[draw,below=0.2cm of anno43, align=center, minimum height = \imageWidth, minimum width = \imageWidth] (anno45){\Large{45}};
    \node[draw,below=0.2cm of anno45, align=center, minimum height = \imageWidth, minimum width = \imageWidth] (anno46){\Large{46}};
    \node[draw,below=0.2cm of anno46, align=center, minimum height = \imageWidth, minimum width = \imageWidth] (anno47){\Large{47}};
    \node[draw,below=0.2cm of anno47, align=center, minimum height = \imageWidth, minimum width = \imageWidth] (anno48){\Large{48}};

    \node[draw,right=3.5cm of anno37, align=center, minimum height = \imageWidth, minimum width = \imageWidth] (anno49){\Large{49}};
    \node[draw,below=0.2cm of anno49, align=center, minimum height = \imageWidth, minimum width = \imageWidth] (anno50){\Large{50}};
    \node[draw,below=0.2cm of anno50, align=center, minimum height = \imageWidth, minimum width = \imageWidth] (anno52){\Large{52}};
    \node[draw,below=0.2cm of anno52, align=center, minimum height = \imageWidth, minimum width = \imageWidth] (anno53){\Large{53}};
    \node[draw,below=0.2cm of anno53, align=center, minimum height = \imageWidth, minimum width = \imageWidth] (anno55){\Large{55}};
    \node[draw,below=0.2cm of anno55, align=center, minimum height = \imageWidth, minimum width = \imageWidth] (anno57){\Large{57}};
    \node[draw,below=0.2cm of anno57, align=center, minimum height = \imageWidth, minimum width = \imageWidth] (anno58){\Large{58}};
    \node[draw,below=0.2cm of anno58, align=center, minimum height = \imageWidth, minimum width = \imageWidth] (anno59){\Large{59}};
    
    \node[draw,right=3.5cm of anno49, align=center, minimum height = \imageWidth, minimum width = \imageWidth] (anno60){\Large{60}};
    \node[draw,below=0.2cm of anno60, align=center, minimum height = \imageWidth, minimum width = \imageWidth] (anno62){\Large{62}};
    \node[draw,below=0.2cm of anno62, align=center, minimum height = \imageWidth, minimum width = \imageWidth] (anno63){\Large{63}};
    \node[draw,below=0.2cm of anno63, align=center, minimum height = \imageWidth, minimum width = \imageWidth] (anno64){\Large{64}};
    \node[draw,below=0.2cm of anno64, align=center, minimum height = \imageWidth, minimum width = \imageWidth] (anno68){\Large{68}};
    \node[draw,below=0.2cm of anno68, align=center, minimum height = \imageWidth, minimum width = \imageWidth] (anno69){\Large{69}};
    \node[draw,below=0.2cm of anno69, align=center, minimum height = \imageWidth, minimum width = \imageWidth] (anno70){\Large{70}};
    \node[draw,below=0.2cm of anno70, align=center, minimum height = \imageWidth, minimum width = \imageWidth] (anno73){\Large{73}};
    
    \node[draw,right=3.5cm of anno60, align=center, minimum height = \imageWidth, minimum width = \imageWidth] (anno74){\Large{74}};
    \node[draw,below=0.2cm of anno74, align=center, minimum height = \imageWidth, minimum width = \imageWidth] (anno80){\Large{80}};
    \node[draw,below=0.2cm of anno80, align=center, minimum height = \imageWidth, minimum width = \imageWidth] (anno88){\Large{88}};
    \node[draw,below=0.2cm of anno88, align=center, minimum height = \imageWidth, minimum width = \imageWidth] (anno89){\Large{89}};
    \node[draw,below=0.2cm of anno89, align=center, minimum height = \imageWidth, minimum width = \imageWidth] (anno98){\Large{98}};
    \node[draw,below=0.2cm of anno98, align=center, minimum height = \imageWidth, minimum width = \imageWidth] (anno99){\Large{99}};
    \node[draw,below=0.2cm of anno99, align=center, minimum height = \imageWidth, minimum width = \imageWidth] (anno100){\Large{100}};
    \node[draw,below=0.2cm of anno100, align=center, minimum height = \imageWidth, minimum width = \imageWidth] (anno101){\Large{101}};

    \foreach \i in \annoVals {
            \node [image,right=of anno\i] (anno_\i_0) {\includegraphics[width=\imageWidth]{imgs/flowers_example/anno/\i_0.png}};
            \node [image,right=of anno_\i_0] (anno_\i_1) {\includegraphics[width=\imageWidth]{imgs/flowers_example/anno/\i_1.png}};
            \node [image,right=of anno_\i_1] (anno_\i_2) {\includegraphics[width=\imageWidth]{imgs/flowers_example/anno/\i_2.png}};
        }

    \node[yshift = -1.2cm, inner sep=0] (text1) at ($(anno15)!0.5!(anno_101_2)$) {\LARGE{Classes in training set}};

    \def\unannoVals{0,1,2,3,8,9,12,13,16,17,18,21,24,27,28,29,31,32,33,35,36,38,39,40,44,51,54,56,61,65,66,67,71,72,75,76,77,78,79,81,82,83,84,85,86,87,90,91,92,93,94,95,96,97}

    \node[draw,below=2.0cm of anno15, align=center, minimum height = \imageWidth, minimum width = \imageWidth] (unanno0){\Large{0}};
    \node[draw,below=0.2cm of unanno0, align=center, minimum height = \imageWidth, minimum width = \imageWidth] (unanno1){\Large{1}};
    \node[draw,below=0.2cm of unanno1, align=center, minimum height = \imageWidth, minimum width = \imageWidth] (unanno2){\Large{2}};
    \node[draw,below=0.2cm of unanno2, align=center, minimum height = \imageWidth, minimum width = \imageWidth] (unanno3){\Large{3}};
    \node[draw,below=0.2cm of unanno3, align=center, minimum height = \imageWidth, minimum width = \imageWidth] (unanno8){\Large{8}};
    \node[draw,below=0.2cm of unanno8, align=center, minimum height = \imageWidth, minimum width = \imageWidth] (unanno9){\Large{9}};
    \node[draw,below=0.2cm of unanno9, align=center, minimum height = \imageWidth, minimum width = \imageWidth] (unanno12){\Large{12}};
    \node[draw,below=0.2cm of unanno12, align=center, minimum height = \imageWidth, minimum width = \imageWidth] (unanno13){\Large{13}};
    \node[draw,below=0.2cm of unanno13, align=center, minimum height = \imageWidth, minimum width = \imageWidth] (unanno16){\Large{16}};
    
    \node[draw,right=3.5cm of unanno0, align=center, minimum height = \imageWidth, minimum width = \imageWidth] (unanno17){\Large{17}};
    \node[draw,below=0.2cm of unanno17, align=center, minimum height = \imageWidth, minimum width = \imageWidth] (unanno18){\Large{18}};
    \node[draw,below=0.2cm of unanno18, align=center, minimum height = \imageWidth, minimum width = \imageWidth] (unanno21){\Large{21}};
    \node[draw,below=0.2cm of unanno21, align=center, minimum height = \imageWidth, minimum width = \imageWidth] (unanno24){\Large{24}};
    \node[draw,below=0.2cm of unanno24, align=center, minimum height = \imageWidth, minimum width = \imageWidth] (unanno27){\Large{27}};
    \node[draw,below=0.2cm of unanno27, align=center, minimum height = \imageWidth, minimum width = \imageWidth] (unanno28){\Large{28}};
    \node[draw,below=0.2cm of unanno28, align=center, minimum height = \imageWidth, minimum width = \imageWidth] (unanno29){\Large{29}};
    \node[draw,below=0.2cm of unanno29, align=center, minimum height = \imageWidth, minimum width = \imageWidth] (unanno31){\Large{31}};
    \node[draw,below=0.2cm of unanno31, align=center, minimum height = \imageWidth, minimum width = \imageWidth] (unanno32){\Large{32}};
    
    \node[draw,right=3.5cm of unanno17, align=center, minimum height = \imageWidth, minimum width = \imageWidth] (unanno33){\Large{33}};
    \node[draw,below=0.2cm of unanno33, align=center, minimum height = \imageWidth, minimum width = \imageWidth] (unanno35){\Large{35}};
    \node[draw,below=0.2cm of unanno35, align=center, minimum height = \imageWidth, minimum width = \imageWidth] (unanno36){\Large{36}};
    \node[draw,below=0.2cm of unanno36, align=center, minimum height = \imageWidth, minimum width = \imageWidth] (unanno38){\Large{38}};
    \node[draw,below=0.2cm of unanno38, align=center, minimum height = \imageWidth, minimum width = \imageWidth] (unanno39){\Large{39}};
    \node[draw,below=0.2cm of unanno39, align=center, minimum height = \imageWidth, minimum width = \imageWidth] (unanno40){\Large{40}};
    \node[draw,below=0.2cm of unanno40, align=center, minimum height = \imageWidth, minimum width = \imageWidth] (unanno44){\Large{44}};
    \node[draw,below=0.2cm of unanno44, align=center, minimum height = \imageWidth, minimum width = \imageWidth] (unanno51){\Large{51}};
    \node[draw,below=0.2cm of unanno51, align=center, minimum height = \imageWidth, minimum width = \imageWidth] (unanno54){\Large{54}};
    
    \node[draw,right=3.5cm of unanno33, align=center, minimum height = \imageWidth, minimum width = \imageWidth] (unanno56){\Large{56}};
    \node[draw,below=0.2cm of unanno56, align=center, minimum height = \imageWidth, minimum width = \imageWidth] (unanno61){\Large{61}};
    \node[draw,below=0.2cm of unanno61, align=center, minimum height = \imageWidth, minimum width = \imageWidth] (unanno65){\Large{65}};
    \node[draw,below=0.2cm of unanno65, align=center, minimum height = \imageWidth, minimum width = \imageWidth] (unanno66){\Large{66}};
    \node[draw,below=0.2cm of unanno66, align=center, minimum height = \imageWidth, minimum width = \imageWidth] (unanno67){\Large{67}};
    \node[draw,below=0.2cm of unanno67, align=center, minimum height = \imageWidth, minimum width = \imageWidth] (unanno71){\Large{71}};
    \node[draw,below=0.2cm of unanno71, align=center, minimum height = \imageWidth, minimum width = \imageWidth] (unanno72){\Large{72}};
    \node[draw,below=0.2cm of unanno72, align=center, minimum height = \imageWidth, minimum width = \imageWidth] (unanno75){\Large{75}};
    \node[draw,below=0.2cm of unanno75, align=center, minimum height = \imageWidth, minimum width = \imageWidth] (unanno76){\Large{76}};
    
    \node[draw,right=3.5cm of unanno56, align=center, minimum height = \imageWidth, minimum width = \imageWidth] (unanno77){\Large{77}};
    \node[draw,below=0.2cm of unanno77, align=center, minimum height = \imageWidth, minimum width = \imageWidth] (unanno78){\Large{78}};
    \node[draw,below=0.2cm of unanno78, align=center, minimum height = \imageWidth, minimum width = \imageWidth] (unanno79){\Large{79}};
    \node[draw,below=0.2cm of unanno79, align=center, minimum height = \imageWidth, minimum width = \imageWidth] (unanno81){\Large{81}};
    \node[draw,below=0.2cm of unanno81, align=center, minimum height = \imageWidth, minimum width = \imageWidth] (unanno82){\Large{82}};
    \node[draw,below=0.2cm of unanno82, align=center, minimum height = \imageWidth, minimum width = \imageWidth] (unanno83){\Large{83}};
    \node[draw,below=0.2cm of unanno83, align=center, minimum height = \imageWidth, minimum width = \imageWidth] (unanno84){\Large{84}};
    \node[draw,below=0.2cm of unanno84, align=center, minimum height = \imageWidth, minimum width = \imageWidth] (unanno85){\Large{85}};
    \node[draw,below=0.2cm of unanno85, align=center, minimum height = \imageWidth, minimum width = \imageWidth] (unanno86){\Large{86}};
    
    \node[draw,right=3.5cm of unanno77, align=center, minimum height = \imageWidth, minimum width = \imageWidth] (unanno87){\Large{87}};
    \node[draw,below=0.2cm of unanno87, align=center, minimum height = \imageWidth, minimum width = \imageWidth] (unanno90){\Large{90}};
    \node[draw,below=0.2cm of unanno90, align=center, minimum height = \imageWidth, minimum width = \imageWidth] (unanno91){\Large{91}};
    \node[draw,below=0.2cm of unanno91, align=center, minimum height = \imageWidth, minimum width = \imageWidth] (unanno92){\Large{92}};
    \node[draw,below=0.2cm of unanno92, align=center, minimum height = \imageWidth, minimum width = \imageWidth] (unanno93){\Large{93}};
    \node[draw,below=0.2cm of unanno93, align=center, minimum height = \imageWidth, minimum width = \imageWidth] (unanno94){\Large{94}};
    \node[draw,below=0.2cm of unanno94, align=center, minimum height = \imageWidth, minimum width = \imageWidth] (unanno95){\Large{95}};
    \node[draw,below=0.2cm of unanno95, align=center, minimum height = \imageWidth, minimum width = \imageWidth] (unanno96){\Large{96}};
    \node[draw,below=0.2cm of unanno96, align=center, minimum height = \imageWidth, minimum width = \imageWidth] (unanno97){\Large{97}};

    \foreach \i in \unannoVals {
            \node [image,right=of unanno\i] (unanno_\i_0) {\includegraphics[width=\imageWidth]{imgs/flowers_example/unanno/\i_0.png}};
            \node [image,right=of unanno_\i_0] (unanno_\i_1) {\includegraphics[width=\imageWidth]{imgs/flowers_example/unanno/\i_1.png}};
            \node [image,right=of unanno_\i_1] (unanno_\i_2) {\includegraphics[width=\imageWidth]{imgs/flowers_example/unanno/\i_2.png}};
        }

    \node[yshift = -1.2cm, inner sep=0] (text1) at ($(unanno16)!0.5!(unanno_97_2)$) {\LARGE{Classes excluded from training set}};

    \end{tikzpicture}}
    \caption{The class splits of the flower dataset, visualized with the class number and three examples for each class.}
    \label{fig:supp_flowers_splits}
\end{figure*}

In \cref{fig:supp_flowers_splits}, we illustrate the split of the flower dataset, featuring classes included in the training data at the top and manually excluded classes (only used during generation) at the bottom. For each class, three example images are provided to convey an impression of the classes.
As detailed in \cref{sub_sec:datasets}, we manually excluded classes primarily showcasing the colors blue, purple, and pink, based on visual inspection of ten examples per class. This split aims to assess our model's ability to adapt to unseen colors.
It is worth noting that some flower types may exhibit a variety of colors, as seen in class 40. Therefore, there is a possibility that some flowers with the colors blue, purple, and pink are included in the training set. However, such cases would be significantly underrepresented, allowing us to still evaluate the benefits of our method.

\section{Classifier-free guidance}
\label{sec:cfg_examples}

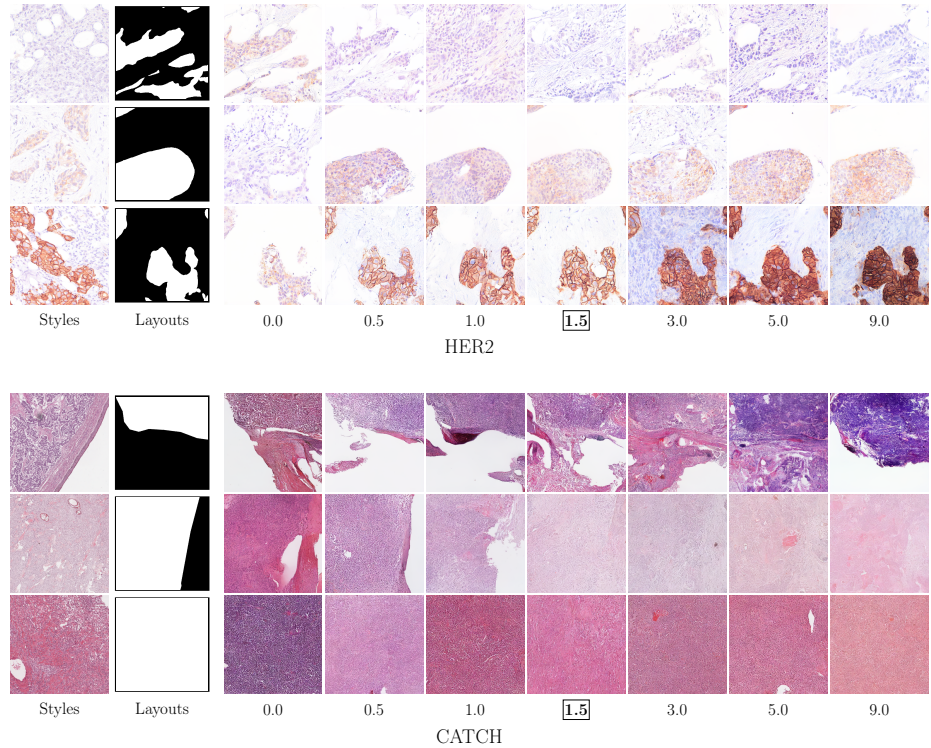
\begin{figure*}
    \centering
    \resizebox{1.0\textwidth}{!}{%
    \begin{tikzpicture}[ image/.style = {inner sep=0pt, outer sep=0pt}, node distance = 1mm and 1mm]
    \def\imageWidth{3cm}
    \def\maskWidth{2.8cm}
    
    \node [image] (frame1) {\includegraphics[width=\imageWidth]{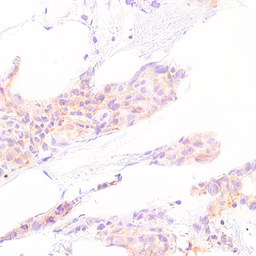}};
    \node [image,right=of frame1] (frame2) {\includegraphics[width=\imageWidth]{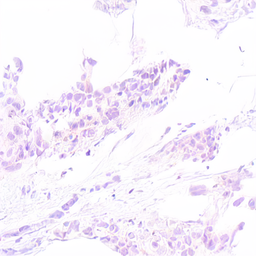}};
    \node [image,right=of frame2] (frame3) {\includegraphics[width=\imageWidth]{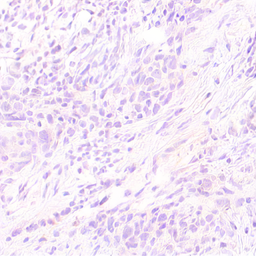}};
    \node [image,right=of frame3] (frame4) {\includegraphics[width=\imageWidth]{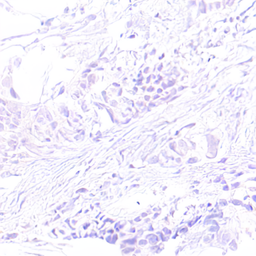}};
    \node [image,right=of frame4] (frame5) {\includegraphics[width=\imageWidth]{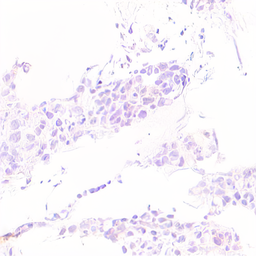}};
    \node [image,right=of frame5] (frame6) {\includegraphics[width=\imageWidth]{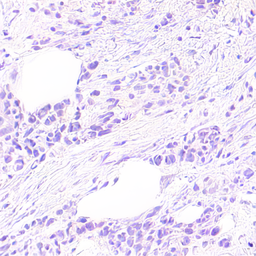}};
    \node [image,right=of frame6] (frame7) {\includegraphics[width=\imageWidth]{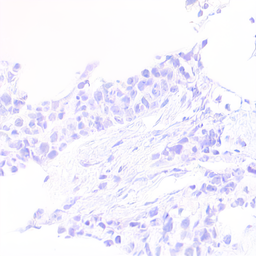}};

    \node[image,below=of frame1] (frame8) {\includegraphics[width=\imageWidth]{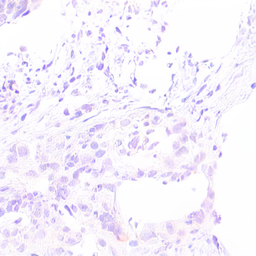}};
    \node[image,right=of frame8] (frame9) {\includegraphics[width=\imageWidth]{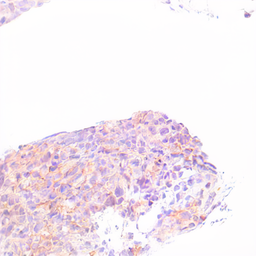}};
    \node[image,right=of frame9] (frame10) {\includegraphics[width=\imageWidth]{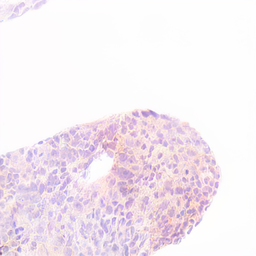}};
    \node[image,right=of frame10] (frame11) {\includegraphics[width=\imageWidth]{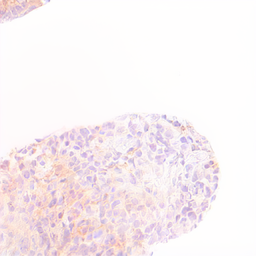}};
    \node[image,right=of frame11] (frame12) {\includegraphics[width=\imageWidth]{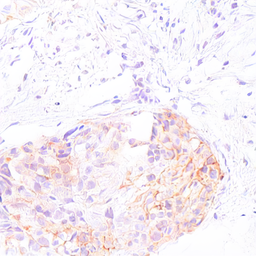}};
    \node[image,right=of frame12] (frame13) {\includegraphics[width=\imageWidth]{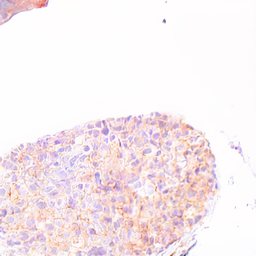}};
    \node[image,right=of frame13] (frame14) {\includegraphics[width=\imageWidth]{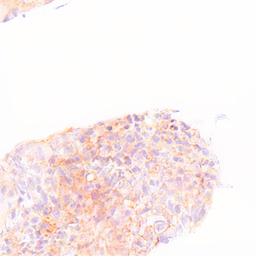}};

    \node[image,below=of frame8] (frame15) {\includegraphics[width=\imageWidth]{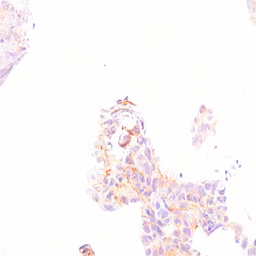}};
    \node[image,right=of frame15] (frame16) {\includegraphics[width=\imageWidth]{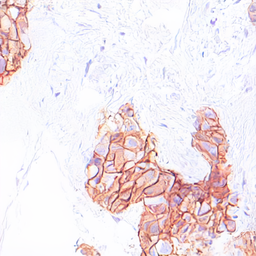}};
    \node[image,right=of frame16] (frame17) {\includegraphics[width=\imageWidth]{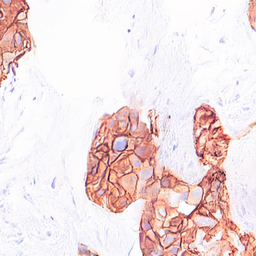}};
    \node[image,right=of frame17] (frame18) {\includegraphics[width=\imageWidth]{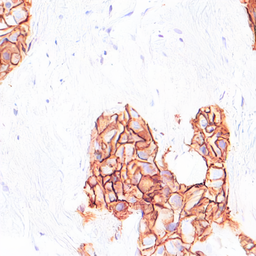}};
    \node[image,right=of frame18] (frame19) {\includegraphics[width=\imageWidth]{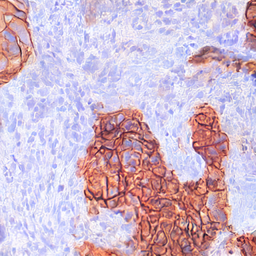}};
    \node[image,right=of frame19] (frame20) {\includegraphics[width=\imageWidth]{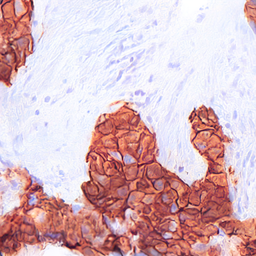}};
    \node[image,right=of frame20] (frame21) {\includegraphics[width=\imageWidth]{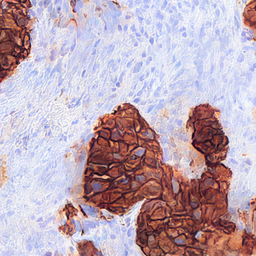}};

    \node[image,left=0.5cm of frame1,draw=black, line width=0.1cm] (style1) {\includegraphics[width=\maskWidth]{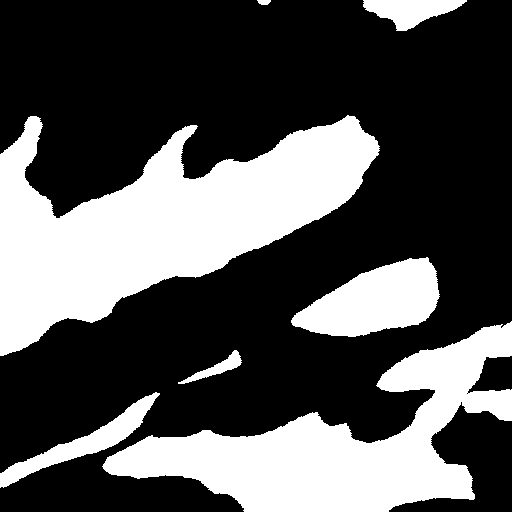}};
    \node[image,left=0.5cm of frame8,draw=black, line width=0.1cm] (style2) {\includegraphics[width=\maskWidth]{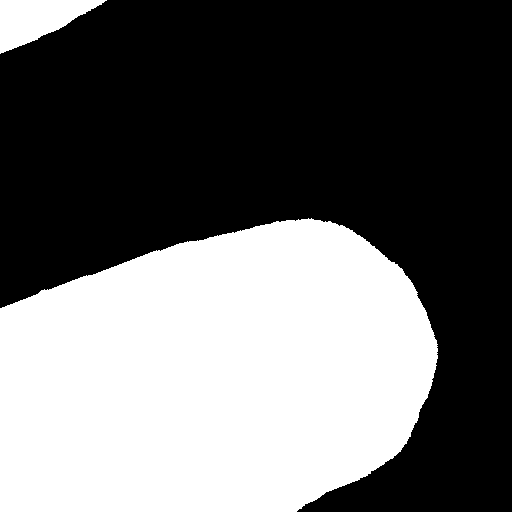}};
    \node[image,left=0.5cm of frame15,draw=black, line width=0.1cm] (style3) {\includegraphics[width=\maskWidth]{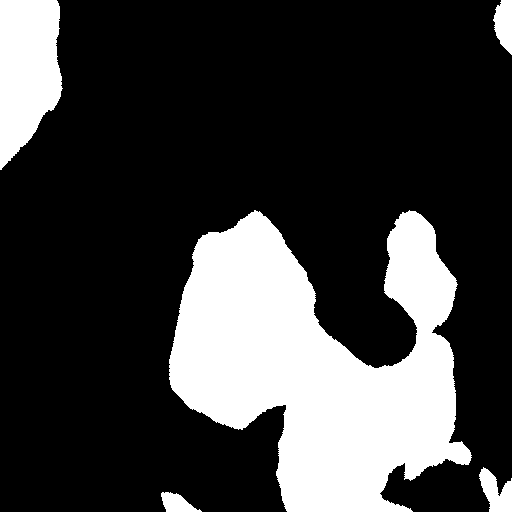}};

    \node[image,left=0.25cm of style1] (mask1) {\includegraphics[width=\imageWidth]{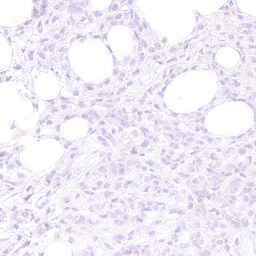}};
    \node[image,left=0.25cm of style2] (mask2) {\includegraphics[width=\imageWidth]{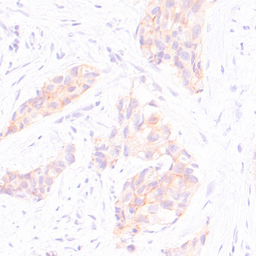}};
    \node[image,left=0.25cm of style3] (mask3) {\includegraphics[width=\imageWidth]{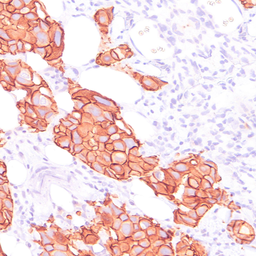}};

    \node [image, below=2.75cm of frame15] (c_frame1) {\includegraphics[width=\imageWidth]{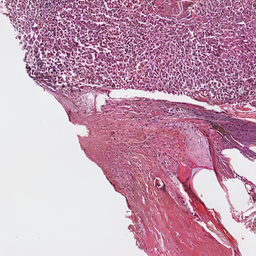}};
    \node [image,right=of c_frame1] (c_frame2) {\includegraphics[width=\imageWidth]{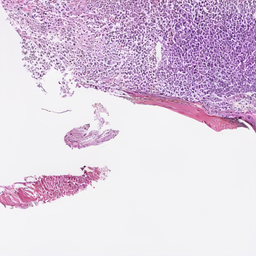}};
    \node [image,right=of c_frame2] (c_frame3) {\includegraphics[width=\imageWidth]{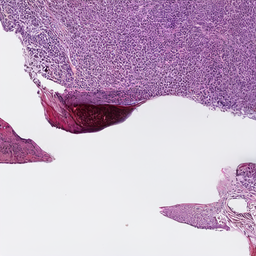}};
    \node [image,right=of c_frame3] (c_frame4) {\includegraphics[width=\imageWidth]{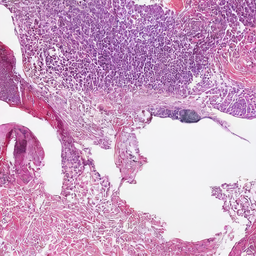}};
    \node [image,right=of c_frame4] (c_frame5) {\includegraphics[width=\imageWidth]{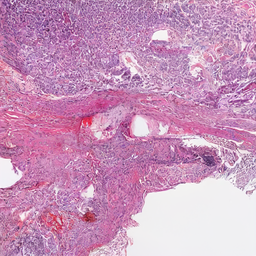}};
    \node [image,right=of c_frame5] (c_frame6) {\includegraphics[width=\imageWidth]{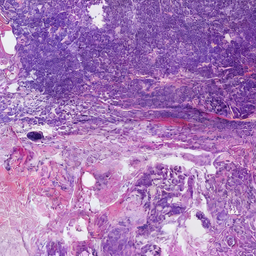}};
    \node [image,right=of c_frame6] (c_frame7) {\includegraphics[width=\imageWidth]{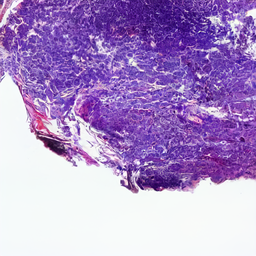}};

    \node[image,below=of c_frame1] (c_frame8) {\includegraphics[width=\imageWidth]{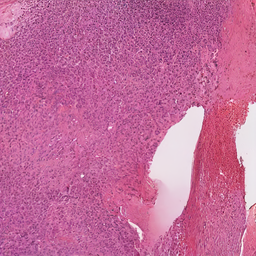}};
    \node[image,right=of c_frame8] (c_frame9) {\includegraphics[width=\imageWidth]{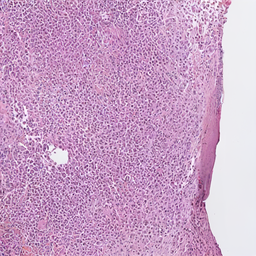}};
    \node[image,right=of c_frame9] (c_frame10) {\includegraphics[width=\imageWidth]{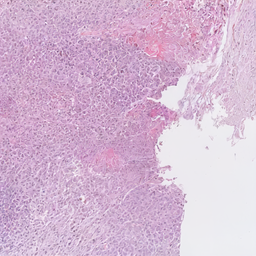}};
    \node[image,right=of c_frame10] (c_frame11) {\includegraphics[width=\imageWidth]{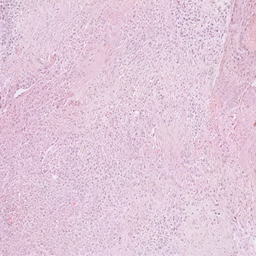}};
    \node[image,right=of c_frame11] (c_frame12) {\includegraphics[width=\imageWidth]{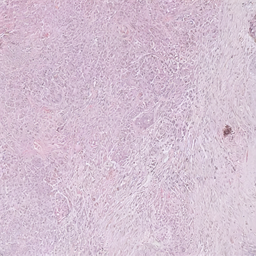}};
    \node[image,right=of c_frame12] (c_frame13) {\includegraphics[width=\imageWidth]{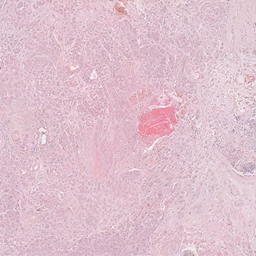}};
    \node[image,right=of c_frame13] (c_frame14) {\includegraphics[width=\imageWidth]{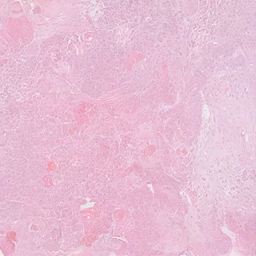}};

    \node[image,below=of c_frame8] (c_frame15) {\includegraphics[width=\imageWidth]{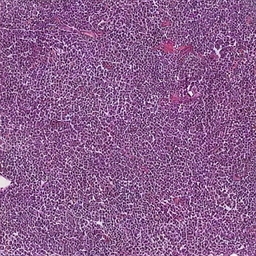}};
    \node[image,right=of c_frame15] (c_frame16) {\includegraphics[width=\imageWidth]{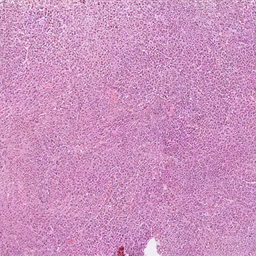}};
    \node[image,right=of c_frame16] (c_frame17) {\includegraphics[width=\imageWidth]{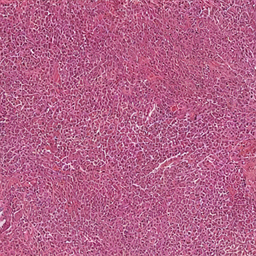}};
    \node[image,right=of c_frame17] (c_frame18) {\includegraphics[width=\imageWidth]{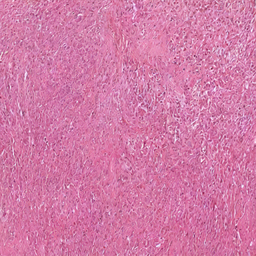}};
    \node[image,right=of c_frame18] (c_frame19) {\includegraphics[width=\imageWidth]{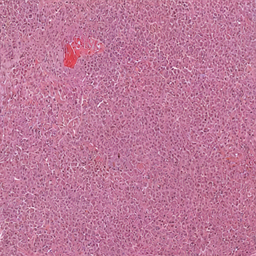}};
    \node[image,right=of c_frame19] (c_frame20) {\includegraphics[width=\imageWidth]{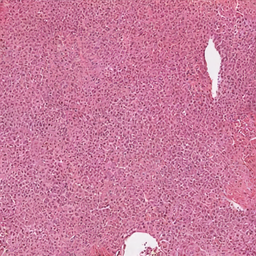}};
    \node[image,right=of c_frame20] (c_frame21) {\includegraphics[width=\imageWidth]{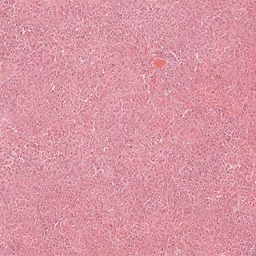}};

    \node[image,left=0.5cm of c_frame1,draw=black, line width=0.1cm] (c_style1) {\includegraphics[width=\maskWidth]{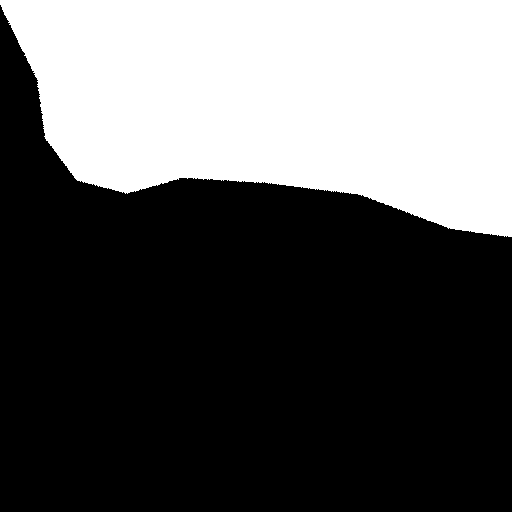}};
    \node[image,left=0.5cm of c_frame8,draw=black, line width=0.1cm] (c_style2) {\includegraphics[width=\maskWidth]{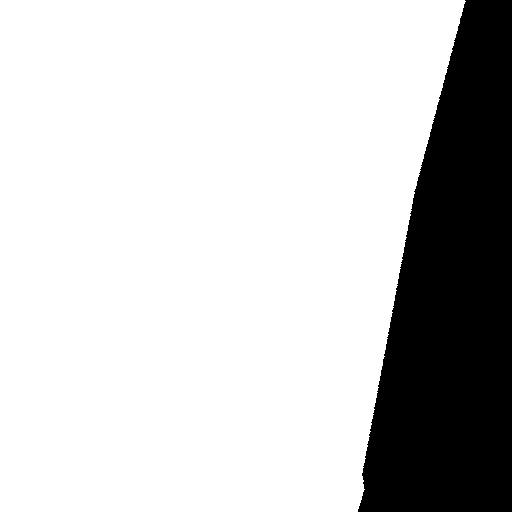}};
    \node[image,left=0.5cm of c_frame15,draw=black, line width=0.1cm] (c_style3) {\includegraphics[width=\maskWidth]{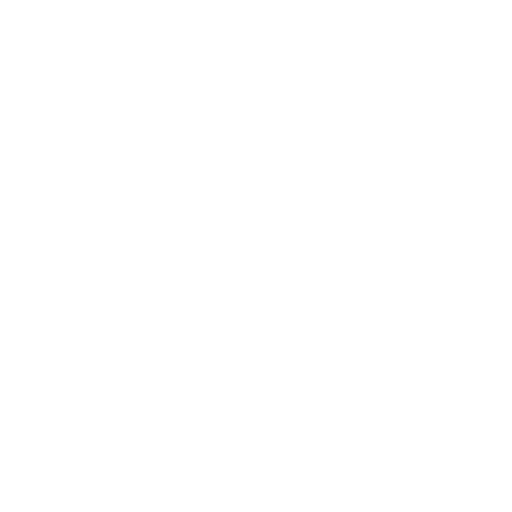}};
    
    \node[image,left=0.25cm of c_style1] (c_mask1) {\includegraphics[width=\imageWidth]{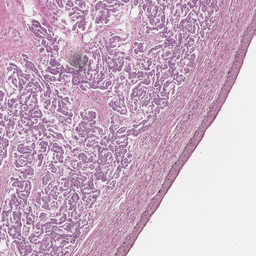}};
    \node[image,left=0.25cm of c_style2] (c_mask2) {\includegraphics[width=\imageWidth]{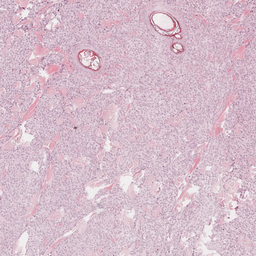}};
    \node[image,left=0.25cm of c_style3] (c_mask3) {\includegraphics[width=\imageWidth]{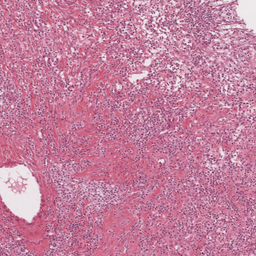}};

    \node[yshift = -2.0cm, inner sep=0] (text1) at ($(style3)$) {\Large{Layouts}};
    \node[yshift = -2.0cm, inner sep=0] (text2) at ($(mask3)$) {\Large{Styles}};

    \node[yshift = -2.0cm, inner sep=0] (text3) at ($(frame15)$) {\Large{0.0}};
    \node[yshift = -2.0cm, inner sep=0] (text4) at ($(frame16)$) {\Large{0.5}};
    \node[yshift = -2.0cm, inner sep=0] (text5) at ($(frame17)$) {\Large{1.0}};
    \node[yshift = -2.0cm, inner sep=0, draw=black, align=center, minimum height=0.6cm, minimum width=0.8cm] (text6) at ($(frame18)$) {\textbf{\Large{1.5}}};
    \node[yshift = -2.0cm, inner sep=0] (text7) at ($(frame19)$) {\Large{3.0}};
    \node[yshift = -2.0cm, inner sep=0] (text8) at ($(frame20)$) {\Large{5.0}};
    \node[yshift = -2.0cm, inner sep=0] (text9) at ($(frame21)$) {\Large{9.0}};

    \node[yshift = -2.8cm] (text3) at ($(mask3)!0.5!(frame21)$) {\LARGE{HER2}};

    \node[yshift = -2.0cm, inner sep=0] (text1) at ($(c_style3)$) {\Large{Layouts}};
    \node[yshift = -2.0cm, inner sep=0] (text2) at ($(c_mask3)$) {\Large{Styles}};

    \node[yshift = -2.0cm, inner sep=0] (text3) at ($(c_frame15)$) {\Large{0.0}};
    \node[yshift = -2.0cm, inner sep=0] (text4) at ($(c_frame16)$) {\Large{0.5}};
    \node[yshift = -2.0cm, inner sep=0] (text5) at ($(c_frame17)$) {\Large{1.0}};
    \node[yshift = -2.0cm, inner sep=0, draw=black, align=center, minimum height=0.6cm, minimum width=0.8cm] (text6) at ($(c_frame18)$) {\textbf{\Large{1.5}}};
    \node[yshift = -2.0cm, inner sep=0] (text7) at ($(c_frame19)$) {\Large{3.0}};
    \node[yshift = -2.0cm, inner sep=0] (text8) at ($(c_frame20)$) {\Large{5.0}};
    \node[yshift = -2.0cm, inner sep=0] (text9) at ($(c_frame21)$) {\Large{9.0}};

    \node[yshift = -2.8cm] (text3) at ($(c_mask3)!0.5!(c_frame21)$) {\LARGE{CATCH}};

    \end{tikzpicture}}
    \caption{Image generation results for the histopathological dataset with nearby style sampling and  classifier-free guidance scales of 0.0, 0.5, 1.0, 1.5, 3.0, 5.0 and 9.0. For our work, we chose a classifier-free guidance scale of 1.5.}
    \label{fig:supp_cfg}
\end{figure*}

In the training phase and during the image generation process, we employed classifier-free guidance for the style query images. This approach proves beneficial as the model learns the style-unconditional distribution of the training data for the omitted style query images, and it also extracts valuable style information when style query images are provided.

For image generation where the requested style lies outside the training style distribution, we utilized classifier-free guidance to compel the model to produce samples beyond the training style distribution. The classifier-free guidance scale determines how far we push the reconstructed image away from the learned style distribution.

Examples of generated images for the histopathological datasets under different classifier-free guidance scales are presented in \cref{fig:supp_cfg}. Lower classifier-free guidance scales yield less style-accurate generated images, while higher scales sometimes result in oversaturated images. Based on visual assessment of the generated images, we selected a classifier-free guidance scale of 1.5 for our experiments, demonstrating accurate styles without oversaturation.

\section{Generations with missing style information}
\label{sec:missing_style}

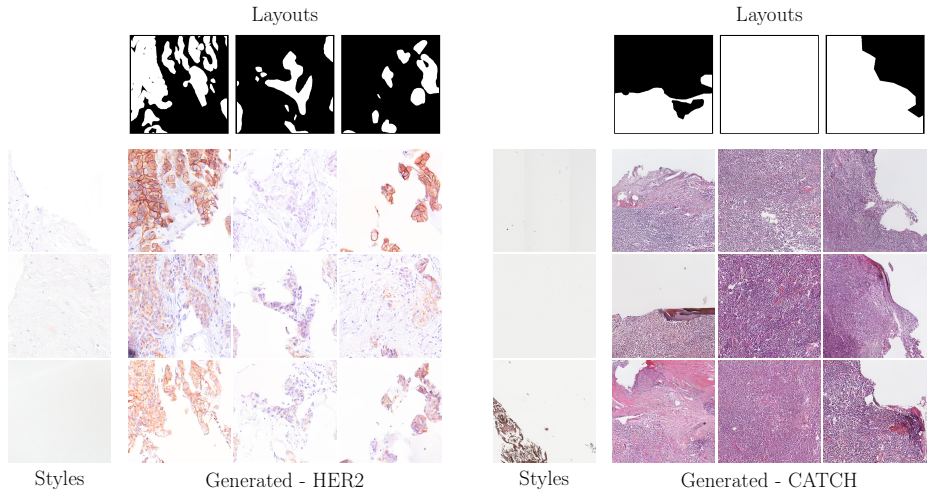
\begin{figure*}
    \centering
    \resizebox{1.0\textwidth}{!}{%
    \begin{tikzpicture}[ image/.style = {inner sep=0pt, outer sep=0pt}, node distance = 1mm and 1mm]
    \def\imageWidth{3cm}
    \def\maskWidth{2.8cm}
    
    \node [image] (frame1) {\includegraphics[width=\imageWidth]{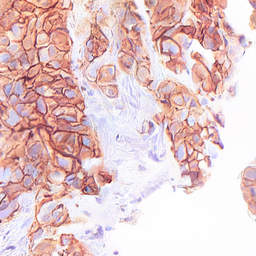}};
    \node [image,right=of frame1] (frame2) {\includegraphics[width=\imageWidth]{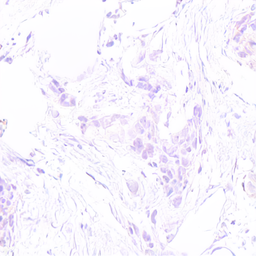}};
    \node [image,right=of frame2] (frame3) {\includegraphics[width=\imageWidth]{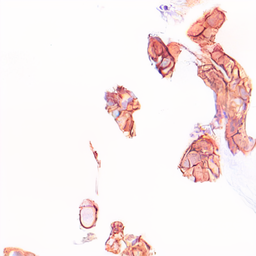}};

    \node[image,below=of frame1] (frame4) {\includegraphics[width=\imageWidth]{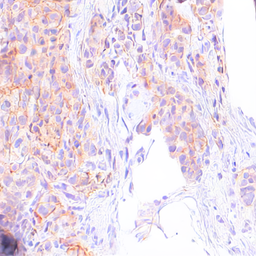}};
    \node[image,right=of frame4] (frame5) {\includegraphics[width=\imageWidth]{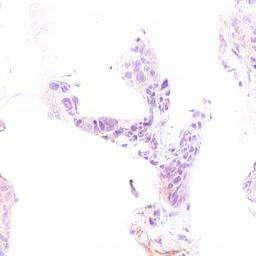}};
    \node[image,right=of frame5] (frame6) {\includegraphics[width=\imageWidth]{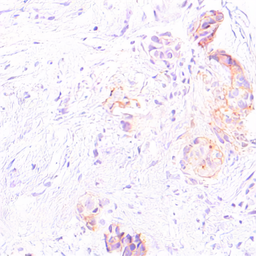}};

    \node[image,below=of frame4] (frame7) {\includegraphics[width=\imageWidth]{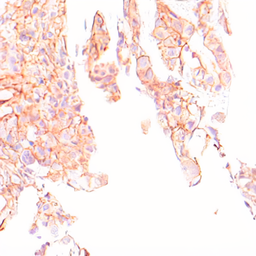}};
    \node[image,right=of frame7] (frame8) {\includegraphics[width=\imageWidth]{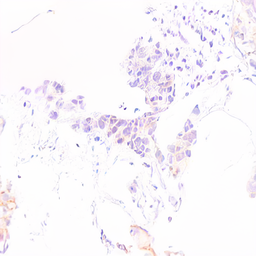}};
    \node[image,right=of frame8] (frame9) {\includegraphics[width=\imageWidth]{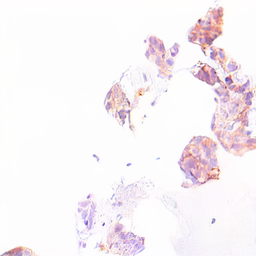}};

    \node[image,left=0.5cm of frame1] (mask1) {\includegraphics[width=\imageWidth]{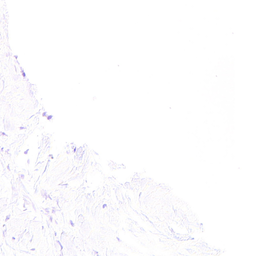}};
    \node[image,left=0.5cm of frame4] (mask2) {\includegraphics[width=\imageWidth]{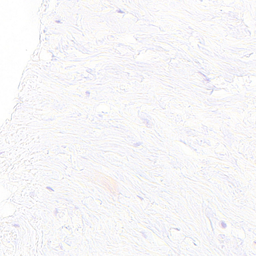}};
    \node[image,left=0.5cm of frame7] (mask3) {\includegraphics[width=\imageWidth]{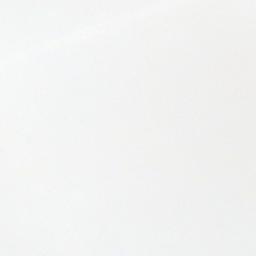}};

    \node[image,above=0.5cm of frame1,draw=black, line width=0.1cm] (style1) {\includegraphics[width=\maskWidth]{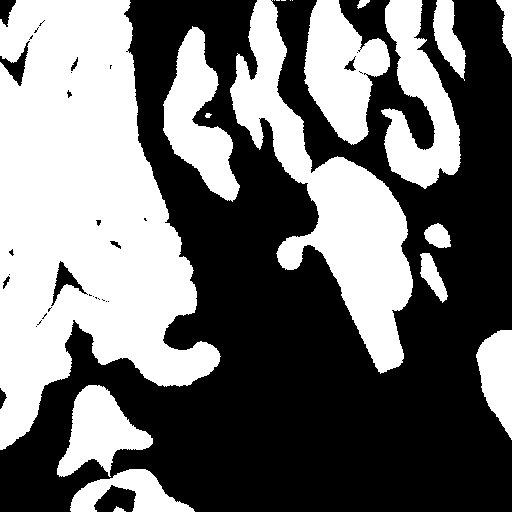}};
    \node[image,above=0.5cm of frame2,draw=black, line width=0.1cm] (style2) {\includegraphics[width=\maskWidth]{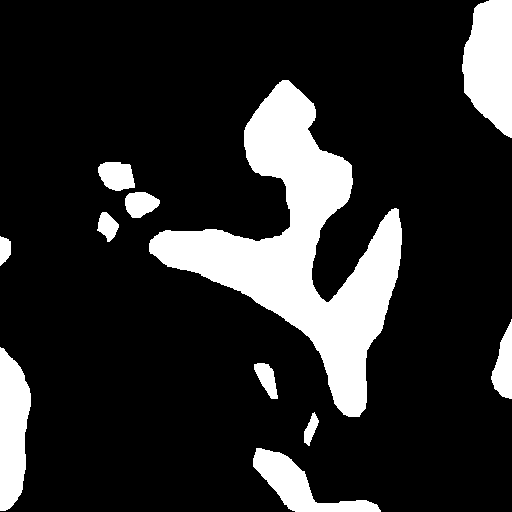}};
    \node[image,above=0.5cm of frame3,draw=black, line width=0.1cm] (style3) {\includegraphics[width=\maskWidth]{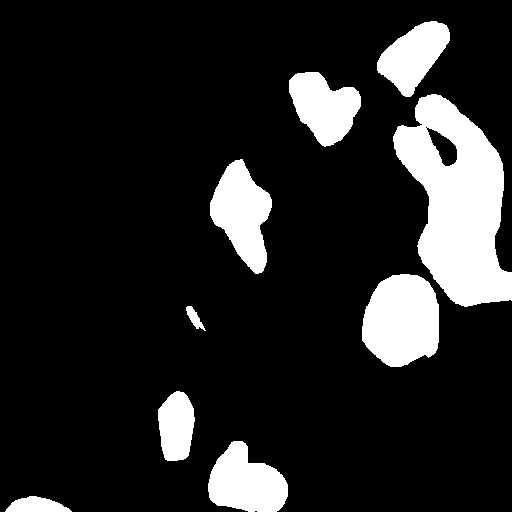}};

    \node [image,right=5cm of frame3] (frame10) {\includegraphics[width=\imageWidth]{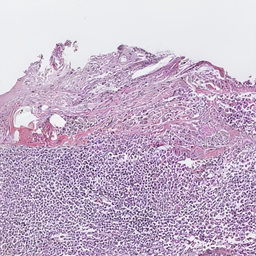}};
    \node [image,right=of frame10] (frame11) {\includegraphics[width=\imageWidth]{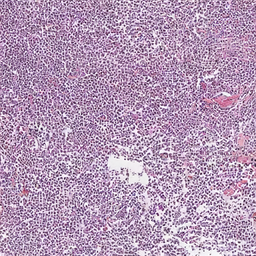}};
    \node [image,right=of frame11] (frame12) {\includegraphics[width=\imageWidth]{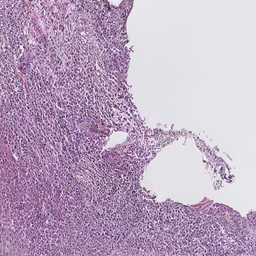}};

    \node[image,below=of frame10] (frame13) {\includegraphics[width=\imageWidth]{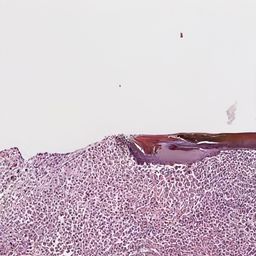}};
    \node[image,right=of frame13] (frame14) {\includegraphics[width=\imageWidth]{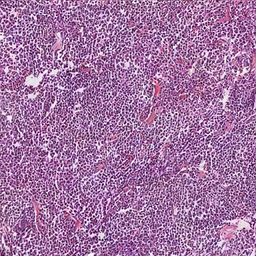}};
    \node[image,right=of frame14] (frame15) {\includegraphics[width=\imageWidth]{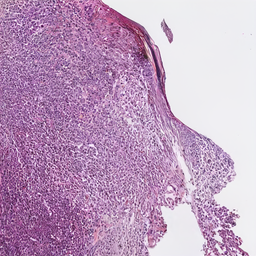}};

    \node[image,below=of frame13] (frame16) {\includegraphics[width=\imageWidth]{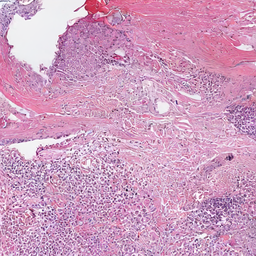}};
    \node[image,right=of frame16] (frame17) {\includegraphics[width=\imageWidth]{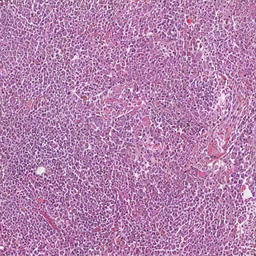}};
    \node[image,right=of frame17] (frame18) {\includegraphics[width=\imageWidth]{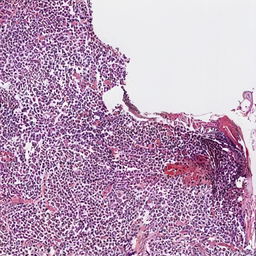}};

    \node[image,left=0.5cm of frame10] (mask4) {\includegraphics[width=\imageWidth]{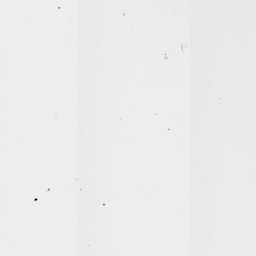}};
    \node[image,left=0.5cm of frame13] (mask5) {\includegraphics[width=\imageWidth]{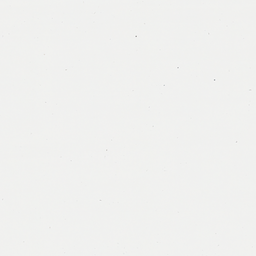}};
    \node[image,left=0.5cm of frame16] (mask6) {\includegraphics[width=\imageWidth]{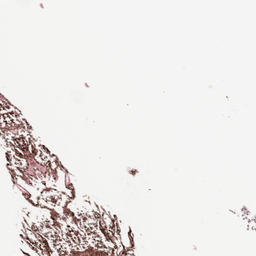}};

    \node[image,above=0.5cm of frame10,draw=black, line width=0.1cm] (style4) {\includegraphics[width=\maskWidth]{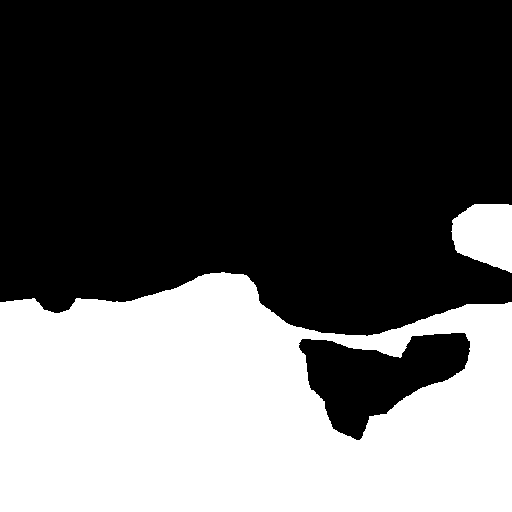}};
    \node[image,above=0.5cm of frame11,draw=black, line width=0.1cm] (style5) {\includegraphics[width=\maskWidth]{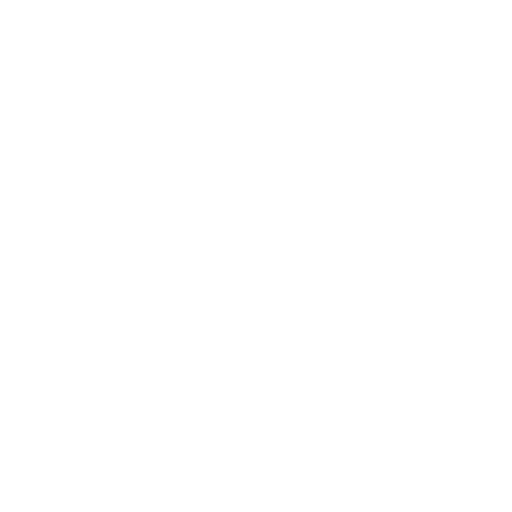}};
    \node[image,above=0.5cm of frame12,draw=black, line width=0.1cm] (style6) {\includegraphics[width=\maskWidth]{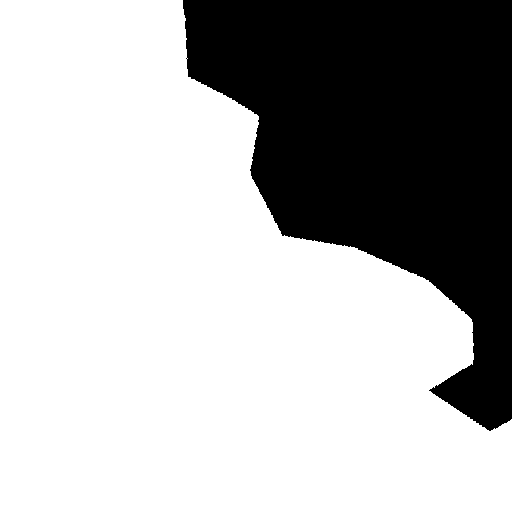}};

    \node[yshift = -2.0cm, inner sep=0] (text1) at ($(frame7)!0.5!(frame9)$) {\LARGE{Generated - HER2}};
    \node[yshift = -2.0cm, inner sep=0] (text2) at ($(frame16)!0.5!(frame18)$) {\LARGE{Generated - CATCH}};
    \node[inner sep=0] (text2) at (mask3|- text1) {\LARGE{Styles}};
    \node[inner sep=0] (text2) at (mask6|- text2) {\LARGE{Styles}};
    \node[yshift = 2.0cm] (text3) at ($(style1)!0.5!(style3)$) {\LARGE{Layouts}};
    \node[yshift = 2.0cm] (text4) at ($(style4)!0.5!(style6)$) {\LARGE{Layouts}};
    \end{tikzpicture}}
    \caption{Image generation results for the histopathological dataset with nearby style sampling. Shown are cases where tumor is present in the layout (white: tumor, black: non-tumor tissue and slide background), but no tumor tissue is present in the style images.}
    \label{fig:supp_missing}
\end{figure*}

In \cref{fig:supp_missing}, we provide examples of image generation for the histopathological datasets in cases where style information is missing in the style query images. The generated images exhibit realistic tissues, but the style is determined by the model and reflects styles from the training style distribution.

This scenario primarily arises in nearby style sampling, where only a single style query image is used. We see these cases as noncritical as they do not result in invalid images, and they highlight that the model falls back to plausible  styles if it cannot extract style information from the style query images.

To ensure that the model respects the style information of valid style query images and does not recreate known styles, we incorporate classifier-free guidance, as discussed in \cref{sec:cfg_examples}.

\section{Flower segmentation results}
\label{sec:flower_seg}

\begin{table*}[]
\resizebox{\textwidth}{!}{%
\begin{tabular}{clccccccccccc}
\toprule
 & Synthetic data  & Mean IoU && IoU Variance && Mean IoU && IoU Variance && Mean IoU && IoU Variance \\
\cmidrule{3-3}\cmidrule{5-5}\cmidrule{7-7}\cmidrule{9-9}\cmidrule{11-11}\cmidrule{13-13}
&& \multicolumn{3}{c}{960 Images} && \multicolumn{3}{c}{480 Images} && \multicolumn{3}{c}{144 Images} \\
\cmidrule{3-5}\cmidrule{7-9}\cmidrule{11-13}
\multicolumn{1}{c}{\multirow{5}{*}{\rotatebox[origin=c]{90}{\textbf{Flowers}}}} &  None               & \textbf{87.80 (0.06)}                 && 4.20 (0.12)                      && \textbf{87.79 (0.26)}                && \textbf{4.14 (0.06)}                      && \textbf{87.05 (0.18)}                && \textbf{4.41 (0.17)}                       \\
& Style Transfer     & 86.67 (0.30)                 && 4.29 (0.17)                      &&                             &&                                  &&                              &&                                   \\
& Semantic DM        & 87.26 (0.17)                 && \textbf{4.18 (0.07)}                      && 87.41 (0.09)               && 4.21 (0.15)                       && 85.91 (0.17)                 && 5.23 (0.25)                       \\
& Augmented (ours)      & 86.49 (0.40)        && 4.48 (0.29)              && 86.26 (0.17)                 && 4.32 (0.08)                       && 85.83 (0.21)                 && 4.86 (0.33)                       \\
\cmidrule{3-13}
\end{tabular}
}
\caption{Segmentation results for the flower dataset, with different amounts of training data and synthetic images.}
\label{table:supp_flower_seg}
\end{table*}

The segmentation results for the flower dataset are presented in \cref{table:supp_flower_seg}. Across all setups, the optimal results were attained when training without synthetic data, with mean IoU scores consistently exceeding 87. The introduction of synthetic images into the training data did not yield improvements in mean IoU scores, although all reported scores remained at high levels, with none dropping below 85. No clear trend in IoU variance between images was evident across the experiments.

We argue that the lack of benefit from synthetic data in the flower dataset is attributable to the task's simplicity, as evidenced by the high IoU scores even at lower/lowest data settings. The ImageNet-pretrained encoder of our segmentation UNet appears capable of adapting to the segmentation task without necessitating the additional information provided by synthetic images. 
Additionally, we argue that the diffusion models could overfit to the layouts, due to the limited number of training examples and the distinct shapes of some flower types, leading to less diverse generated images. For the histopathological datasets, this problem does not exist, since even for low amounts of data, no connection between images and layouts exists.

\end{document}